\documentclass{article}

\PassOptionsToPackage{numbers, compress}{natbib}
\usepackage[final]{neurips_data_2021}

\usepackage[utf8]{inputenc}
\usepackage[T1]{fontenc}
\usepackage{hyperref}
\usepackage{url}
\usepackage{booktabs}
\usepackage{amsfonts}
\usepackage{nicefrac}
\usepackage{microtype}
\usepackage{xcolor}
\usepackage{multirow}
\usepackage{graphicx}
\usepackage{subcaption}
\usepackage{float}
\usepackage{placeins}

\hypersetup{
  pdftitle={An Extensible Benchmark Suite for Learning to Simulate Physical Systems},
  pdfauthor={Karl Otness, Arvi Gjoka, Joan Bruna, Daniele Panozzo, Benjamin Peherstorfer, Teseo Schneider, Denis Zorin},
}

\input{macros.tex}

\title{An Extensible Benchmark Suite for Learning to Simulate Physical Systems}

\author{%
  Karl Otness\\
  Courant Institute of Mathematical Sciences\\
  New York University\\
  \texttt{karl.otness@nyu.edu}\\
  \And
  Arvi Gjoka\\
  Courant Institute of Mathematical Sciences\\
  New York University\\
  \texttt{arvi.gjoka@nyu.edu}\\
  \And
  Joan Bruna\\
  Courant Institute of Mathematical Sciences\\
  New York University\\
  \And
  Daniele Panozzo\\
  Courant Institute of Mathematical Sciences\\
  New York University\\
  \And
  Benjamin Peherstorfer\\
  Courant Institute of Mathematical Sciences\\
  New York University\\
  \And
  Teseo Schneider\\
  University of Victoria\\
  \hphantom{Courant Institute of Mathematical Sciences}\\
  \And
  Denis Zorin\\
  Courant Institute of Mathematical Sciences\\
  New York University\\
}

\makeatletter
\let\neuripsnotice\@noticestring
\renewcommand{\@noticestring}{\phantom{\neuripsnotice}}
\makeatother

\begin{document}

\maketitle

\begin{abstract}
  Simulating physical systems is a core component of scientific computing,
  encompassing a wide range of physical domains and applications.
  Recently, there has been a surge in data-driven methods to complement traditional numerical simulations methods, motivated by the
  opportunity to reduce computational costs and/or learn new
  physical models leveraging access to large collections of data.
  However, the diversity of problem settings and applications has led to a
  plethora of approaches, each one evaluated on a different setup and
  with different evaluation metrics.
  We introduce a set of benchmark problems to take a step towards unified benchmarks and evaluation
  protocols. We propose four representative physical systems, as
  well as a collection of both widely used classical time
  integrators and representative data-driven methods (kernel-based,
  MLP, CNN, nearest neighbors). Our framework allows evaluating
  objectively and systematically the stability, accuracy, and
  computational efficiency of data-driven methods. Additionally, it is configurable to permit
  adjustments for accommodating other learning tasks and for establishing a foundation
  for future developments in machine learning for scientific computing.
\end{abstract}

\section{Introduction}%
\label{sec:introduction}

Computational modeling of physical systems 
is a core task of scientific computing.  Standard methods
rely on discretizations of explicit models typically given in the form of partial differential equations (PDEs). Machine learning techniques can extend these techniques in a number of ways. In some cases, a closed system of analytic equations relating all variables may not be available (e.g., a constitutive relation for a material may not be known). In other cases, while a full analytic description of a system is available, a traditional solution may be too costly (e.g., turbulence) or can be sped up substantially using data-driven reduced-order models.
However, despite promising results, a successful adoption of these data-driven approaches into scientific computing pipelines requires a solid and exhaustive assessment of their performance---a challenging task given the diversity of physical systems, corresponding data-driven approaches, and the lack of standardized sets of problems, comparison protocols, and metrics.

We focus on the setting where the physical model is unavailable during training, mimicking situations in computational science and engineering with ample data and a lack of models. One can generally distinguish two 
different flavors of physical simulation with different associated computational cost: those that map a high-dimensional state space into another high-dimensional space (as in temporal integration schemes, mapping the state of the system at one time step to the next), or from a high-dimensional input space to a lower-dimensional output (as in surrogate models, mapping the initial conditions to a functional of the solution). 
While this distinction also applies to data-driven approaches, another critical aspect emerges from the choice of input data distribution. 
We identify two extremes:
the \emph{narrow} data regime, where initial conditions are sampled from a low-dimensional manifold (even within a high-dimensional state space), and the \emph{wide} regime, where initial conditions span a truly high-dimensional space. As could be expected, narrow data regimes define an easier prediction task where data-driven methods can potentially `bypass the physics', whereas wide regimes require models with enough encoded physical priors in order to beat the curse of dimensionality. Therefore, such choice of data distribution is a critical component of any data-driven physical simulation benchmark. 

In this work, we introduce an extensible benchmark suite, including:
\textbf{(1)} an extensible set of simple, yet representative, physical models with a range of training and evaluation (test) setups, as well as reference, high-accuracy numerical solutions to benchmark data-driven methods, \textbf{(2)} reference implementations of traditional time integration schemes, which are used as baselines for evaluation, and \textbf{(3)} implementations of widely used data-driven methods, including  physics-agnostic multi-layer perceptrons (MLPs), convolutional neural networks (CNNs),
kernel machines and non-parametric nearest neighbors. 
Our benchmark suite is modular, permitting extensions with limited code changes, and captures both `narrow' and `wide' regimes by appropriately parametrizing the set of initial conditions. 

Our analysis reveals two important conclusions. First, even in the simplest physical models, current data-driven pipelines, while providing qualitatively acceptable solutions, are quantitatively far from directly numerically integrating physical models, and this performance gap appears unfeasible to close by merely scaling up the models and/or the dataset size.
In other words, the cost of ignoring the physics is high, even for the simplest physics, and cannot in general be compensated by data, matching insights that have been obtained in other scientific computing settings~\cite{doi:10.1098/rsta.2016.0153,Willcox2021}.
Next, and more importantly, our simple $L^2$-based nearest neighbor regressor is used to calibrate how `narrow' the learning task is.
Our finding is that even for seemingly complex systems, such as the incompressible Navier-Stokes systems, such naive predictor outperforms most deep-learning-based models in the narrow regime---thus providing a simple calibration of the true difficulty of the simulation task, that we advocate should be present in every future evaluation.  

\section{Related work}%
\label{sec:related-work}

Machine learning is used in physical simulation in a number of interrelated ways. Some important uses include reduced-order/surrogate modeling, learning constitutive models or more generally compact analytic representations from data.
A unifying theme of these applications of machine learning is automatic construction of parametric models capable of reproducing the behavior of physical systems for a sufficiently broad range of initial data, boundary conditions and other system parameters.  The purpose of these representations varies from acceleration (e.g., surrogate machine learning models are used to accelerate optimization), to automatic construction of multiscale models (learning macroscopic constitutive laws from microscopic simulation), to inferring compact descriptions of unknown representations from experimental data.

The purpose of our proposed benchmarks is to enable comparisons of different learning-based methods in terms of their accuracy and efficiency.
We briefly review two streams of learning methods for physical systems: 
\textbf{(1)} One line of work aims to understand how neural networks can be structured and trained to reproduce known physical system behavior, with the goal of designing general methods applicable in a variety of settings \citep{greydanus19,sanchezgonzalez19,chen19,sanchezgonzalez18,raissi2017physics,raissi2019physics,lu2019deepxde,Haghighat:2020,tartakovsky2018learning,thuerey20}.
Our benchmark cases fit primarily into this category.
\textbf{(2)} Another line of research aims to develop a variety of techniques to accelerate solving PDEs.  Typically, these methods are developed for specific PDEs and a specific restricted range of problems: for example, fluid dynamics problems \citep{ribeiro2020deepcfd,kim2019deep,xie2018tempogan}, with particular applications to cardiovascular modeling \citep{liang2020feasibility,kissas2020machine} and aerodynamics \citep{umetani2018learning}; or solid mechanics simulation tasks, including stresses \citep{nie2020stress,Liang:2018,maso2020deep,Khadilkar2019,li2018reconstruction,liaimage}.
In cases where the governing equations are not given, the learning task becomes approximating them from data \citep{Geometry-from-a-time-series,Crutchfield87equationsof,ANTOULAS01011986,772353,InterpMOR,doi:10.1137/16M1106122,Brunton12042016,doi:10.1098/rspa.2016.0446,doi:10.1137/18M116798X,SchmidDMD,FLM:7843190,Tu2014391,QKPW19LiftLearn}.

\section{Background and problem setup}%
\label{sec:background}

\paragraph{PDEs, dynamical systems, and time integration}
Consider a time-dependent PDE of the form $\partial_t u = \mathcal{L}(u)$, where $u$ is the 
unknown function and $\mathcal{L}$ is a possibly nonlinear operator that includes spatial derivatives of $u$. By discretizing in space, one obtains a dynamical system
\begin{equation}
\dot{x}(t) = f(x(t))
\label{eq:sysdef}
\end{equation}
with an $N$-dimensional state $x(t) \in \mathbb{R}^{N}$ at time $t\in[0,T]$. The function $f$ is assumed to be Lipschitz to ensure solution uniqueness and the initial condition is denoted as $x_0 \in \mathbb{R}^N$.
A PDE of a higher order in time can be reduced to the first-order form in the standard way, e.g.,  
if we have a second-order system $\ddot{q}(t) = f(q(t))$, then we consider its formulation via position $q$ and momentum $p$  as a first-order system with $x = [q;p]$:
$[\dot{q}(t); \dot{p}(t)] = [p(t); f(q(t))]$.
To numerically integrate \eqref{eq:sysdef}, we choose time steps $0 = t_0 < t_1 < \dots < t_K = T$. Then, a time integration scheme (e.g., \citep{suli03,HairerI,HairerII}) gives an approximation $x_k \approx x(t_k)$ of the state $x(t_k)$ at each time step $k = 1, \dots, K$. A list of the schemes we use along with details is given in Appendix~\ref{sec:int-detail}.

\paragraph{Problem setup and learning problems}
Given $M$ initial conditions $x^{(1)}_0, \dots, x^{(M)}_0 \in \mathbb{R}^N$ and the corresponding $M$ trajectories $X^{(i)} = [x^{(i)}_0, \dots, x^{(i)}_K] \in \mathbb{R}^{N \times (K + 1)}, i = 1, \dots, M$ obtained with a time integration scheme from dynamical system \eqref{eq:sysdef},
we consider the following two learning problems, both of which aim to learn the physical model of the problem, viewed as unknown, from trajectory samples: 
\textbf{(1)} Learning an approximation $\tilde{f}$ of the right-hand side function $f$ in Eq.~\eqref{eq:sysdef}. This gives an approximate $\dot{\tilde{x}}(t) = \tilde{f}(\tilde{x}(t))$ that is then numerically integrated to produce a trajectory $\tilde{X}$ for an initial condition $\tilde{x}_0$. The aim is that $\tilde{X}$ approximates well the true trajectory $X$ obtained with $f$ from \eqref{eq:sysdef} for the same initial condition.
   \textbf{(2)} Directly learning the next steps in the trajectory from the current one, i.e. predict $x^{(i)}_{k}$ given $x^{(i)}_{k - 1}$.

To assess the learned models, we evaluate them on their ability to produce good approximate trajectories from randomly sampled initial conditions, by either integration or direct step prediction. During evaluation, we use initial conditions drawn independently from those used to produce training data, both from the same distribution as the training samples, as well as from a distribution with support outside the training range. We train networks on data sets of various sizes. For details, see Appendix~\ref{sec:method-detail}.

\section{Benchmark systems}%
\label{sec:sys-detail}

\begin{figure}
\centering
\includegraphics[width=\textwidth]{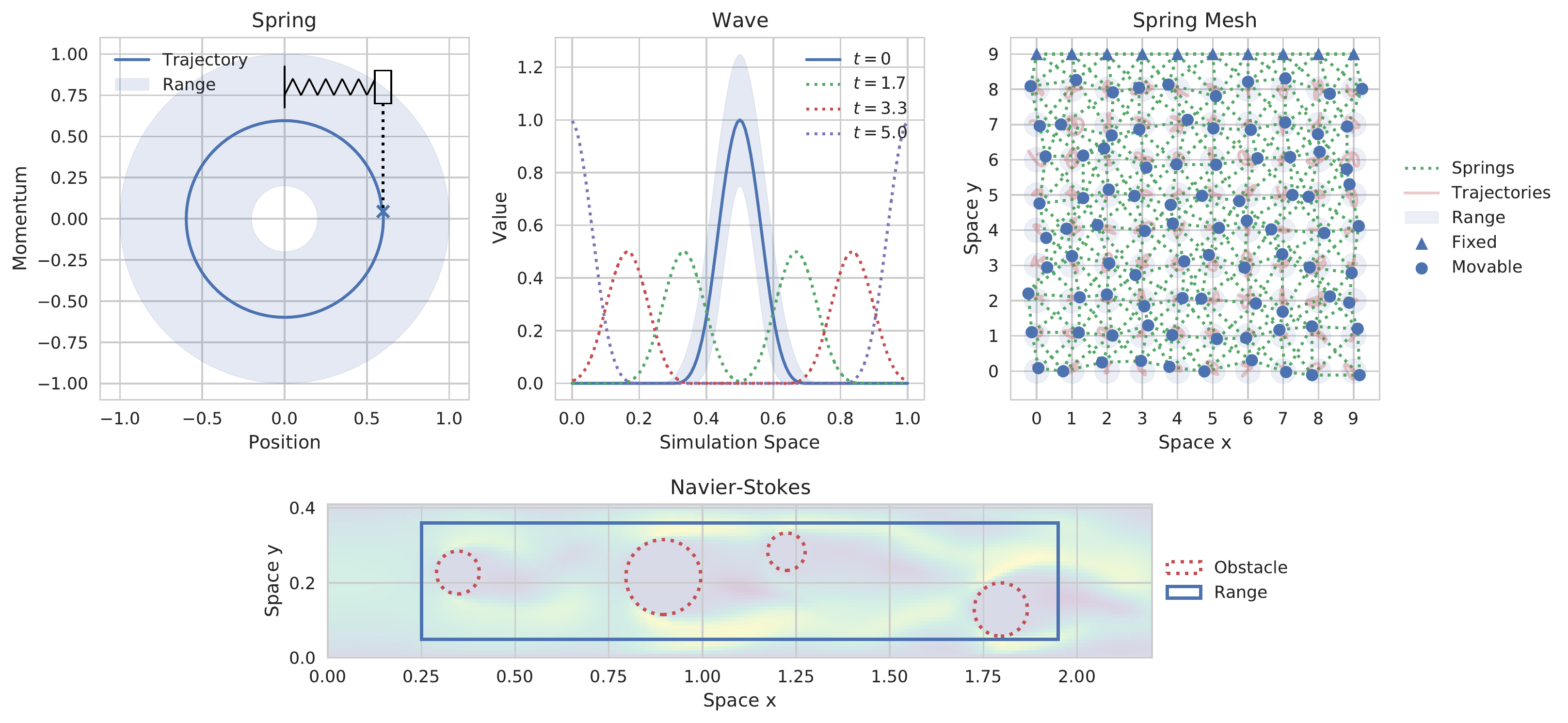}
\caption{Representative visualizations of the four systems, depicting the results and ranges of initial condition sampling. Each has two state components: for the Navier-Stokes system, a flow velocity and a pressure field, and for the other three a position $q$ and momentum $p$.}
\label{fig:sysillust-brief}
\end{figure}

We consider four physical systems, illustrated in Figure \ref{fig:sysillust-brief}: a single oscillating spring, a one-dimensional linear wave equation, a Navier-Stokes flow problem and a mesh of damped springs. 
These systems represent a progression of complexity: the spring system is a linear system with low-dimensional space of initial conditions and low-dimensional state; the wave equation is a low-dimensional linear system with a (relatively) high-dimensional state space after discretization;
the Navier-Stokes equations are nonlinear and we consider a setup with low-dimensional initial conditions and high-dimensional state space; finally, the spring mesh system has both high-dimensional initial conditions as well as high-dimensional states.  Additionally, the proposed spring system and Navier-Stokes problems represent diffusion-dominated and advection-dominated (for sufficiently low viscosity) PDE behaviors, 
as well as variability in initial conditions with fixed domain (spring system) and variable domain (Navier-Stokes). 
These varying complexities provide an opportunity to test methods on simpler systems and the ability to examine changing performance as system size increases, both in terms of the state dimension, and the initial condition distribution.
The ground truth models for the spring, wave, and spring mesh systems with classical time integrators are implemented using NumPy \citep{numpy}, SciPy \citep{scipy}, and accelerated, where possible, with Numba \citep{numba}. The Navier-Stokes snapshots are generated using PolyFEM \citep{polyfem}, a finite element library.

These systems were chosen in an effort to reflect the variety of systems used for testing in this area, while unifying choices of particular formulations. Past works have chosen systems of the types featured here: simple oscillators (both spring and pendulum~\cite{greydanus19}), particle systems with various interaction laws (gravity, spring forces, charges, cloth simulations, etc.~\cite{cranmer20,kipf18,sanchezgonzalez19,chen19,pfaff20}), and fluid-flow systems (with various sorts of obstacles, airfoils or cylinders~\cite{thuerey20,pfaff20}). We make particular selections here in an effort to unify systems of interest and facilitate comparisons across experiments by providing a shared set of tasks which can be used for development and testing of machine learning methods.

Some examples of initial condition selection for each system are illustrated in Figure~\ref{fig:sysillust-brief}. The ground truth for the spring, wave, and spring mesh systems consists of the state variables $(q, p)$ for position and momentum, and their associated derivatives $(\dot{q}, \dot{p})$. For the Navier-Stokes system the state consists of flow velocities, and a pressure field, along with approximated time derivatives for each.

Table~\ref{tab:train-param} lists the parameters used to generate trajectories for training and evaluation. Training sets of three sizes are generated, each containing the specified number of trajectories. The systems are integrated at the listed time step sizes, but the ground truth data is subsampled further by the factor shown after $\div$ in the table: the integration schemes are run at a smaller time step and intermediate computations are discarded. Each larger training set is a strict superset of its predecessor to ensure that previous training samples are never removed.

\subsection{Spring}%
\label{sec:spring}
We simulate a simple one-dimensional oscillating spring. In this system, the spring has zero rest length, and both the oscillating mass and spring constant are set to $1$. The spring then exerts a force inversely proportional to the position of the mass $q$: $\dot{p}(t) = -q$ and $\dot{q}(t) = p$.

The energy of the system is proportional to $r=q^2+p^2$ which is the radius of the circle in phase space. To sample initial conditions, we first sample a radius uniformly, then choose an angle theta uniformly. This produces a uniform distribution over spring system energy levels and starts at an arbitrary point in the cycle. The spring system has a closed-form solution: $(q(t), p(t)) = (r\sin(t+\theta_0), r\cos(t+\theta_0))$ where $r$ is the radius of the circle traced in phase space (the energy of the spring) and $\theta_0$ is the phase space angle at which the oscillation will start. While this closed form solution is useful, for consistency with our other systems, we generate snapshots of the spring system by numerical integration. Simulations of the spring system always run through one period. For ``in-distribution'' training values, the radius is selected in the range $(0.2, 1)$ and ``out-of-distribution'' radii are chosen from $(1, 1.2)$.  

\subsection{Wave}%
\label{sec:wave}

This benchmark system is similar to the one used in \citet{peng16}. Consider the wave equation with speed $c=0.1$
\begin{align}
\partial_{tt} u = c^2\partial_{xx}u\,,
\end{align}
on a one-dimensional spatial domain $[0, 1)$ with periodic boundary conditions. We represent this second-order system as a first-order system and discretize in space to obtain
\begin{align}
\begin{bmatrix}
\dot{q}(t)\\
\dot{p}(t)
\end{bmatrix} = \begin{bmatrix}
0 &  I\\
c^2 D_{xx} & 0
\end{bmatrix}\begin{bmatrix}
q(t)\\
p(t)
\end{bmatrix}\,,
\end{align}
where $D_{xx} \in \mathbb{R}^{n \times n}$ corresponds to the three-point central difference approximation of the spatial derivative $\partial_{xx}$ and the matrices $I$ and $0$ are the identity and zero matrix, respectively, of appropriate size. We discretize in space with $n=125$ evenly spaced grid points and evolve the system following the dynamics described above.

Initial conditions are sampled with an initial pulse in the $q$ component centered at $0.5$. All initial conditions have zero momentum. The initial pulse is produced by a spline kernel as described in \cite{peng16}:
\[
  s(x) = \frac{10}{p_w} \cdot \abs{x - 0.5}\,,\qquad 
  h(s) = p_h \cdot
         \begin{cases}
           1 - \frac32 s^2 + \frac34s^3 & \text{if $0 \leq s \leq 1$}\\
           \frac14(2-s)^3 & \text{if $1 < s \leq 2$}\\
           0 & \text{else}
         \end{cases}
\]
where the width and height of the pulse are scaled by parameters $p_w$ and $p_h$, respectively. The spline kernel pulse is then $h(s(x))$ for $x\in[0,1)$, evaluated at the discretized grid points.

For ``in-distribution'' samples, parameters $p_w, p_h$ are both chosen uniformly in the range $(0.75, 1.25)$ and ``out-of-distribution'' runs sample uniformly from $(0.5, 0.75) \cup (1.25, 1.5)$. All trajectories are integrated until $t=5$ when the wave has traveled through half a period.

\subsection{Spring mesh}%
\label{sec:spring-mesh}
This system manipulates a square grid of particles connected by springs, in a two dimensional space, 
and can be considered a simplified version of deformable surface and volume systems (cf. \cite{pfaff20}).
The particles all have mass $1$, and are arranged into a unit grid. Springs are added along the axis-aligned edges and diagonally across each grid square, with rest lengths selected so that the regularly-spaced particles are in a rest position.

In this work we use a $10\times10$ grid where the top row of particles is fixed in place. Initial conditions are sampled by choosing a perturbation for the position of each non-fixed spring. These perturbations are chosen as uniform vectors inside a circle with radius $0.35$. Out-of-distribution perturbations are chosen uniformly in a ring with inner radius 0.35 and outer radius 0.45. The sampled initial conditions all have zero momentum.

In this system, a spring between particles $a$ and $b$ exerts a force:
\begin{equation}
  F_{ab} = -k \cdot \dparen[\big]{\norm{q_a - q_b}_2 - \ell_{ab}}\frac{q_a - q_b}{\norm{q_a-q_b}_2} - \gamma \dot{q}_a
\end{equation}
where $\ell_{ab}$ is the rest length of the spring, $\gamma=0.1$ is a parameter controlling the magnitude of an underdamped velocity-based decay, and $k=1$ is the spring constant.

\subsection{Navier-Stokes}

We consider the standard Navier-Stokes equation over a domain $\Omega$ (cf. \cite{pfaff20,thuerey20})
\begin{align*}
\left. \begin{array}{l}
\displaystyle\rho\frac{\partial u}{\partial t} + \rho(u \cdot \nabla) u - \nu \Delta u + \nabla p = b \\
\nabla \cdot u = 0 \\
u(0)  = u_0 
\end{array}\right\}\text{ on } \Omega\times(0, T)~,~\left.\begin{array}{l}
u  = d  \\
\nu\frac{\partial u}{\partial n} + pn  = g
\end{array}\right\}\text{ on } \partial\Omega_D\times(0, T) ~,
\end{align*}
where $u\colon \Omega\times(0, T) \to R^2$ is the velocity
at time $t\in (0, T)$ of a fluid with kinematic viscosity $\nu$ and density $\rho$, $p\colon \Omega\times(0, T) \to R$ is the pressure and $\partial\Omega_D$ and $\partial\Omega_N$ are the Dirichlet and Neumann boundary conditions, respectively. In our setup we use the finite element method (FEM) to solve the PDE using mixed discretization: quadratic polynomial for the velocity and linear for pressure. In our experiment the domain $\Omega$ is a rectangle $0.22\times 0.41$ with a randomly generated set of circular obstacles. We start with $u_0=0$ and specify a velocity on the left boundary of $u(0,y) = (6(1-e^{-5t})(0.41-y)y/0.1681, 0)$, zero on the top and bottom, and zero Neumann on the right side ($g=0$). We solve the system using PolyFEM~\cite{polyfem} using $dt=0.08$ and backward differentiation formula (BDF) of order 3 for the time integration.

We sample obstacles into two configurations: a single obstacle, or a set of four. In each case, we sample the obstacles leaving a margin of 0.05 between each circle, and a margin of 0.25 from the left and right sides, and 0.05 between the top and bottom. Otherwise, each obstacle is determined by first sampling a radius, then sampling a center from the valid space, respecting the margins. If the sampled obstacle is too close to an existing circle, it is discarded and a new sample is drawn. In-distribution obstacles have radii in the range $(0.05, 0.1)$ and out-of-distribution radii are drawn from $(0.025, 0.05)$.

\begin{table}
  \caption{Dataset sizes and simulation parameters}
  \label{tab:train-param}
  \centering
  \begin{tabular}{lllll}
    \toprule
    System & \# Train Trajectories & \# Eval Trajectories & Time Step Size & \# Steps\\
    \midrule
    Spring & 10, 500, 1000 & 30 & 0.00781, $\div 128$ & 805\\
    Wave & 10, 25, 50 & 6 & 0.00049, $\div 8$ & 10204\\
    Spring Mesh & 25, 50, 100 & 15 & 0.00781, $\div 128$ & 805\\
    Navier-Stokes & 25, 50, 100 & 5 & 0.08, $\div 1$ & 65\\
    \bottomrule
  \end{tabular}
\end{table}
\section{Numerical experiments}%
\label{sec:results}

\paragraph{Experimental setup}
We apply several basic learning methods to the datasets developed in this work: $k$-nearest neighbor regressors, a neural network kernel method, several sizes of feed-forward MLPs, and a variety of CNNs. 
Details of the architectures and the training protocol are provided in supplementary material, Appendix~\ref{sec:method-detail}. Each of the neural networks we consider is implemented using PyTorch \citep{pytorch}.

The learning methods considered in this work are each trained on one of the two target task formulations described in Section \ref{sec:background}. For  derivative-based prediction, the training is conducted supervised on ground truth snapshots gathered from the underlying models. For each system we randomly sample initial conditions and each of these is then numerically integrated to produce a trajectory. Each trajectory includes state samples $x$ as well as target derivatives $\dot{x}$ used for training. For direct prediction, we no longer require numerical integration; instead we directly predict the trajectory in a sequential fashion. In this setting, we approximate $\tilde{f}_\theta(x(t)) \approx x(t + \delta_t)$ for a discrete time step size $\delta_t$. For the derivative prediction task we report results using the leapfrog integrator. Full results using other numerical integration schemes are available in the supplementary materials.

We pick the same set of learning methods and apply it to both tasks independently to judge performance in each. For many systems the state is divided into position and momentum components: $x \equiv (q, p)$. For the Navier-Stokes problem, the state $x$ is made up of the flow velocity field, and the scalar field for pressure. After training, we produce rolled-out trajectories from held-out initial conditions, either by combining with a numerical integrator in the case of derivative prediction, or in a directly recurrent fashion in the case of step prediction.
Each neural network is instantiated in three independent copies, each of which is trained and evaluated across all sampled trajectories.
We compute a per-step MSE against a ground truth value, average these per-step MSEs to produce a per-trajectory error, and record these errors for analysis. Our experiments are designed to test several aspects of physical simulation. We highlight the most salient ones next, and report more extensive results in Appendix \ref{sec:plots}. 

\begin{figure}
    \centering
    \includegraphics[width=\textwidth]{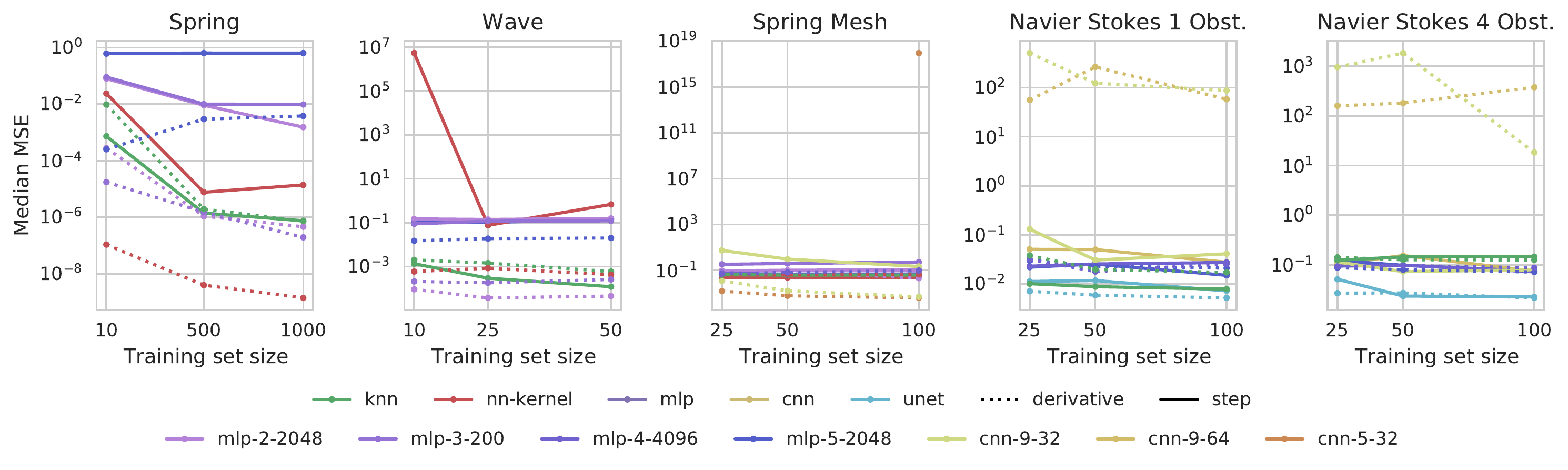}
    \caption{Median MSE error with respect to the training set size for each of our system configurations.
    We show varying architecture choices for each method.}
    \label{fig:experiment_summary1}
\end{figure}

\paragraph{Training set size} In general ML problems, one would expect additional training samples to systematically improve (in-distribution) evaluation performance. However, Figure \ref{fig:experiment_summary1} illustrates a clear saturation of performance on the simplest systems when using neural networks as function approximators, in contrast with non-parametric KNNs and the kernel method. We attribute this saturation to an inherent gap between the training and evaluation objectives. While data-driven methods are optimized to minimise next-step predictions, the final evaluation requires built-in stability to prediction errors. Including regularisation strategies to incorporate stability, such as noise injection \cite{pfaff20}, is shown to help, but not fully resolve this issue. 

\paragraph{Out-of-distribution evaluation} For simplicity, we only examine the out-of-distribution error for networks trained on the largest training set size. The added challenge of out-of-distribution samples varies with the construction of each system. It is possible to get some idea of the difficulty increase by examining the accuracy penalty for the KNNs, and comparing it to how well the more advanced models are able to generalize.

\begin{figure}
    \centering
    \includegraphics[width=\textwidth]{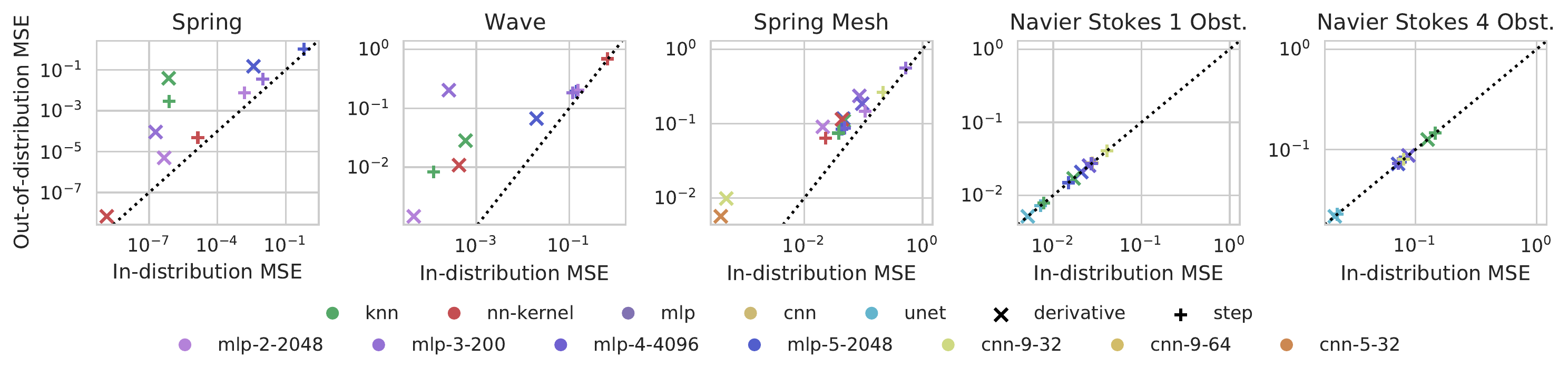}
    \caption{
    Median MSE for in-distribution evaluation sets vs.\ out-of-distribution evaluation sets for each system. Colors represent the same method and are varied for different architecture choices. Marker shapes distinguish step and derivative prediction, and the dotted line is the identity line. Outliers for the spring mesh and both Navier-Stokes configurations were removed. Values on both axes were approximately $10^{13}$ for the spring mesh and in the range $10^2$--$10^3$ for Navier-Stokes.}
    \label{fig:experiment_summary2}
\end{figure}

Benefits of neural networks for generalization over KNN are visible across several systems in Figure~\ref{fig:experiment_summary2}, particularly in the spring system for small MLPs for derivative prediction and nn-kernel in both cases. The KNN suffers a significant increase in error while these methods produce only somewhat worse predictions. Benefits are still present, though less pronounced, for the wave system derivative prediction where neural networks increase in error, but the kernel method and small MLP maintain a lower absolute error than the KNN.
On the Navier-Stokes systems none of the methods suffers an increase in error for out-of-distribution evaluation. The change in radius distribution for the obstacles did not pose an additional challenge sufficient to produce a measurable change in error distribution. We attribute this to low dimension of the initial condition space.

\begin{figure}
    \centering
    \includegraphics[width=\textwidth]{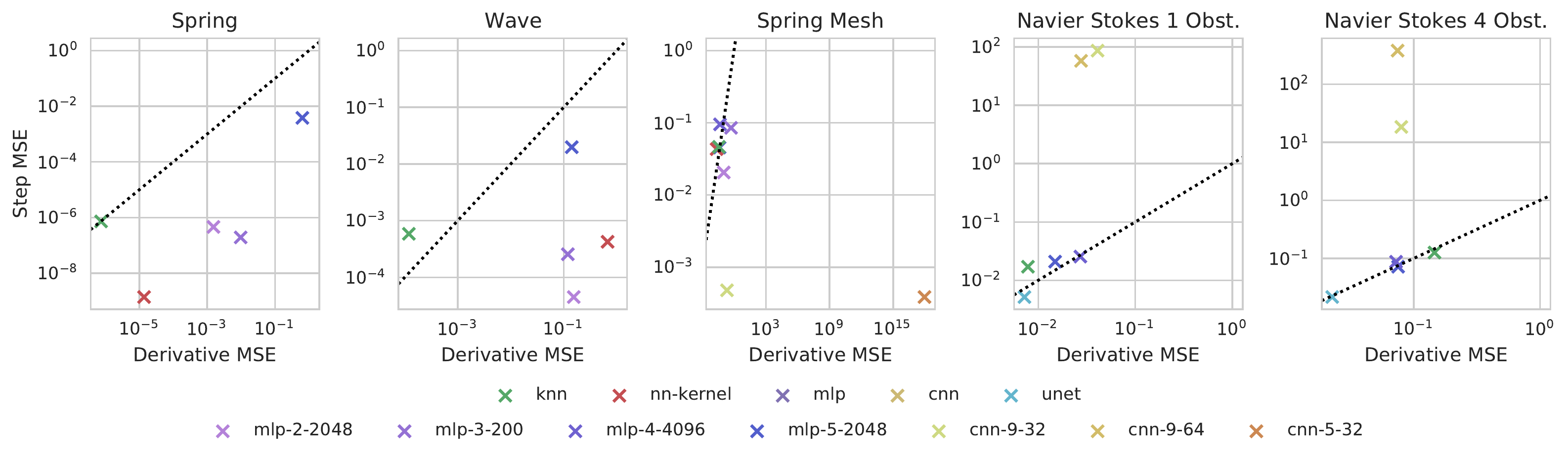}
    \caption{Median MSEs for derivative vs.\ step prediction on the same evaluation set. Results are displayed for each of our system configurations. The dotted line is the identity line.}
    \label{fig:experiment_summary3}
\end{figure}

\paragraph{Step and derivative prediction} 
The step and derivative prediction instances of each learning problem lead to different accuracy from the learning methods we test. While most physical systems are naturally described in terms of their derivatives through corresponding ODEs/PDEs, data-driven simulations also offer the alternative of bypassing this differential formulation and predict the next state directly. Such `cavalier' approach avoids the compounding error amplification effects across integration steps, at the expense of sample efficiency. Figure \ref{fig:experiment_summary3} illustrates these tradeoffs across our systems. 

An important example of this effect is the performance of CNNs on the spring mesh system (Figure~\ref{fig:sm10-error-dist} in the Appendix). When working through a numerical integrator and performing derivative prediction they produce the lowest error of all methods tested, but following the same training protocol for step prediction these architectures produce high errors, or are unstable. This case is likely an interaction of the architecture with the specific learning task. For the spring mesh, step prediction requires outputting the position of the particle which requires manipulating its global coordinates, while derivative predictions allow the network to more easily act locally and compute only a relative motion for the particle. The derivative prediction task better takes advantage of the spatial invariance of the CNNs. This difference in performance reflects the importance of tailoring architectures to the specific task, and some potential for neural network architectures to benefit from incorporating knowledge of a system's behavior.

\paragraph{System and dataset complexity}
Several trends we observe correlate with the difficulty of learning to simulate a system, and the variation in its behavior across the training and evaluation samples. This is generally a combination of the system's state dimension, and variation in its behavior,  approximated by the dimension of the distribution from which initial conditions are sampled.

\begin{figure}
    \centering
    \includegraphics[width=0.85\textwidth]{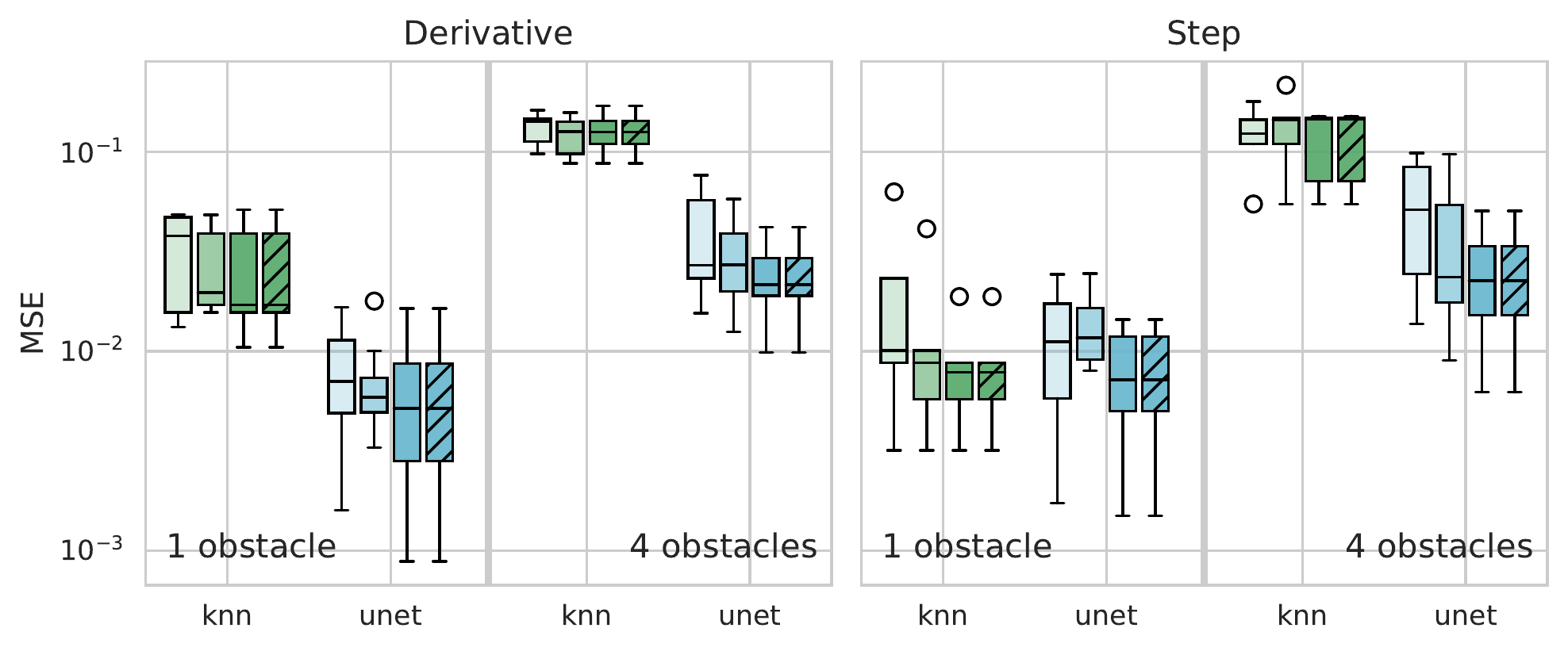}
    \caption{MSE distribution across trajectories in the evaluation set for KNNs and u-nets on our Navier-Stokes system, both the formulation with one obstacle, and the settings with four, as well as for derivative and step prediction. Box plot mid-lines show medians, box area is between first and third quantiles, whiskers extend $1.5\times$ beyond the boxes, outliers are plotted past the whiskers. Darker colors denote increasing training set size. The final hatched box is the same network from the final un-hatched box, tested on an out-of-distribution evaluation set.}
    \label{fig:experiment_summary4}
\end{figure}

This is particularly visible in Figure \ref{fig:experiment_summary4} in the performance of the KNN methods, and, in many cases, the performance of simpler methods such as the small MLPs. On the simpler systems, such as the spring and wave, the KNNs generally perform well because even though the wave system has a relatively large state dimension of 125, like the spring its initial condition is sampled from only two parameters and its behavior can be readily predicted from these. The Navier-Stokes system with a single obstacle is another instance of this sort of behavior: the KNN is readily able to reproduce flows it has not seen because a sampling of 100 obstacle positions is such that an evaluation sample is close to a trajectory seen at training time. Therefore, small MLPs and the kernel method produce similar performance. When the difficulty is increased by sampling four obstacles, the KNN and MLP performances suffer, and larger networks such as the u-net are needed to maintain approximately the same performance.

\paragraph{Choice of numerical integrator}
For our derivative prediction tasks we combine our trained methods with three explicit integrators with orders 1, 2, and 4. In most of our systems these produce at most a small increase in accuracy, holding all other training and evaluation parameters equal. However on the Navier-Stokes system the higher order integrators produce somewhat higher errors, particularly for the u-net and the MLPs. This appears related to the approximated derivatives used for training this system. The learned derivatives produce some small deviations which are compounded when combining multiple derivative samples.

\begin{table}
  \caption{Time comparison for derivative prediction against baseline numerical integrators}
  \label{tab:timing-results}
  \centering
\resizebox{0.92\textwidth}{!}{
\begin{tabular}{llrrrrrr}
\toprule
\multirow{2}{*}{System} & \multirow{2}{*}{Architecture} & \multicolumn{2}{c}{Euler} & \multicolumn{2}{c}{Leapfrog} & \multicolumn{2}{c}{RK4}\\
\cmidrule(lr){3-4} \cmidrule(lr){5-6} \cmidrule(l){7-8}
& & Time Ratio & Scaling & Time Ratio & Scaling & Time Ratio & Scaling\\
\midrule
\multirow{5}{*}{Spring} & knn & 367.0 & 1$\times$ & 405.8 & 16$\times$ & 311.3 & 64$\times$\\
& nn kernel & 180.7 & 1$\times$ & 198.6 & 1$\times$ & 173.3 & 1$\times$\\
& mlp-2-2048 & 185.8 & 1$\times$ & 191.6 & 16$\times$ & 177.7 & 64$\times$\\
& mlp-3-200 & 237.6 & 1$\times$ & 237.5 & 16$\times$ & 227.2 & 64$\times$\\
& mlp-5-2048 & 473.8 & 4$\times$ & 369.6 & 64$\times$ & 360.9 & 128$\times$\\
\addlinespace
\multirow{5}{*}{Wave} & knn & 24,102.7 & 8$\times$ & 16,945.1 & 256$\times$ & 16,132.0 & 256$\times$\\
& nn kernel & 35.3 & 8$\times$ & 22.3 & 256$\times$ & 19.9 & 256$\times$\\
& mlp-2-2048 & 25.4 & 8$\times$ & 16.5 & 256$\times$ & 14.2 & 256$\times$\\
& mlp-3-200 & 31.0 & 16$\times$ & 19.7 & 256$\times$ & 17.8 & 256$\times$\\
& mlp-5-2048 & 60.5 & 16$\times$ & 38.0 & 256$\times$ & 34.6 & 256$\times$\\
\addlinespace
\multirow{8}{*}{Spring Mesh} & knn & 708.6 & 8$\times$ & 626.1 & 128$\times$ & 690.4 & 256$\times$\\
& nn kernel & 5.3 & 8$\times$ & 4.4 & 128$\times$ & 4.8 & 256$\times$\\
& mlp-2-2048 & 3.0 & 8$\times$ & 2.6 & 128$\times$ & 2.8 & 256$\times$\\
& mlp-3-200 & 3.4 & 8$\times$ & 3.2 & 128$\times$ & 3.4 & 256$\times$\\
& mlp-4-4096 & 7.8 & 16$\times$ & 7.1 & 128$\times$ & 7.7 & 256$\times$\\
& mlp-5-2048 & 7.1 & 8$\times$ & 6.2 & 128$\times$ & 5.3 & 256$\times$\\
& cnn-9-32 & 11.0 & 2$\times$ & 10.5 & 32$\times$ & 11.3 & 64$\times$\\
& cnn-5-32 & 7.4 & 1$\times$ & 9.2 & 32$\times$ & 7.3 & 64$\times$\\
\bottomrule
\end{tabular}
}
\end{table}

\paragraph{Computational overheads} Another important aspect to
consider when applying learning methods to physical simulation
problems is the time required to compute each step, and the
computational overheads introduced by the lack of knowledge of the
true system physics. With standard numerical integration methods, it is
generally possible to improve the quality of generated trajectories by
decreasing the size of the time step used during integration. We take
advantage of this in order to estimate the time overheads of our
learning methods relative to our baseline numerical integrators at
approximately corresponding error levels.

We numerically integrate each system at time step sizes scaled by
powers of two. For each trajectory in the derivative prediction
setting, we find the smallest scaling factor at which the numerical
integrator exceeds the learning method's error at their final shared
time step, approximating the factor by which numerical integration can be
made faster until it begins to underperform the learned method.

Table~\ref{tab:timing-results} reports the results of these
experiments. For each numerical integrator, the ``scaling'' column
reports the most common scaling factor found for each trajectory. The
``time ratio'' column represents the learned method's evaluation
overhead (median times,
counting only per-step network evaluation costs, not numerical
integration or data transfers). Note that the numerical integrator
makes fewer steps than the learned method so the overall trajectory
time must be further adjusted by the scaling factor.

In general, the neural networks are slower per-step by one or two
orders of magnitude. KNNs are slower by significantly larger factors,
particularly for the wave system. This is likely partially due to the
default scikit-learn KNN implementation used, and due to the large
size of the wave system training sets (large state dimension and large
number of training snapshots). Scaling factors increase with the order
of the integrator as higher-order integrators are more tolerant of
large step sizes and maintain low error.

It is likely that these overheads could be reduced with more optimized
implementations of both the numerical integrators and learned methods.
The derivative prediction task is also constrained by its need to
interact with the numerical integrator. In this setting the learned
methods cannot be expected to outperform the quality of the solutions
generated by the true system derivatives. This reflects a penalty
resulting from a lack of knowledge of the true underlying system, and
a penalty for learning from observations in this case. Step prediction
without involving the numerical integrator potentially avoids some of
these constraints, if learning is successful.

\section{Conclusions and limitations}%
\label{sec:conclusion}

The results in this work illustrate the performance achievable by applying common machine learning methods to the simulation problems in our proposed benchmark task. We envision three ways in which the results of this work might be used: \textbf{(1)} the datasets developed here can be used for training and evaluating new machine learning techniques in this area, \textbf{(2)} the simulation software can be used to generate new datasets from these systems of different sizes, different initial condition dimensionality and distribution; the training software could be used to assist in conducting further experiments, and \textbf{(3)} some of the trends seen in our results may help inform the design of future machine learning tasks for simulation.

For the first and second groups of downstream users, we have made available the pre-generated datasets used in this work, as well as the software used to produce them and carry out our experiments. These components allow carrying out the measurements we have made here, and permit further adjustments to be made. Documentation on using the datasets and the software is included in Appendix~\ref{sec:dataset-info}.

For the third group, we highlighted a few trends that suggest useful steps to take in developing new problems and datasets in this area. First, we advise including several simple baseline methods when designing new tasks. In particular the inclusion of standard numerical integrators (for derivative-type problems) and KNNs are useful to evaluate the difficulty of the proposed task. Specifically, KNNs are useful for examining the performance achievable by memorizing the training set, and are thus \emph{witnessing} an appropriate design of data distribution that captures the true high-dimensionality of the prediction task. 
As an example, in the Navier-Stokes examples some task formulations may inadvertently be simple to memorize, even if the complexity of the system itself may not immediately suggest it. 
The numerical integrators are likewise useful as baselines both to ensure that the derivative learning is feasible even when achieving no error in predictions, and also to evaluate the penalty in accuracy which is incurred by operating without access to the true physics.
We believe that in light of these observed trends, including baseline methods such as standard numerical integration schemes and simple learning methods such as the KNN is important in understanding tasks in this area. Including these assists in experiment design by helping to calibrate the difficulty of a target task.

\paragraph{Limitations} While our benchmark provides actionable conclusions on a wide array of simulation domains, it is currently focused on temporal integration, and as such it does not cover important settings in scientific computing. For instance, we do not currently include an instance of a surrogate model, which could provide different tradeoffs benefiting ML models. Additionally, we have focused on two setups for data-driven simulation (differential snapshot prediction and direct snapshot prediction), but other alternatives exist that might mitigate some of the shortcomings we observed; for instance by considering larger temporal contexts (as in \cite{chen19}), as well as enforcing certain conservation laws into the model \cite{greydanus19,chen19}.
Finally, while we report some
measurements of timings and relative computational overheads, there
are other dimensions to the time-accuracy tradeoff which remain to be
explored and further software optimizations are most likely possible.

\FloatBarrier

{
  \small
  \bibliography{main,main-ML}
  \bibliographystyle{abbrvnat}
}

\newpage

\appendix
\section{Numerical integration schemes}%
\label{sec:int-detail}

We briefly review the time integration schemes that we consider in this study: forward Euler (FE), leapfrog (LF), Runge-Kutta 4 (RK4), and backward Euler (BE). Other sources also discuss these integration schemes, for example \citet{suli03,HairerI,HairerII}.

Time integration with the explicit Euler method leads to
\[
x_{k} = x_{k - 1} + \deltat f(x_{k - 1}),
\]
where $\deltat > 0$ is the time step size and $f$ is the right-hand side function.
The explicit Runge-Kutta 4 scheme is
\[x_k = x_{k - 1} + \frac{\deltat}{6}\left(h_1 + 2h_2 + 2h_3 + h_4\right),\]
where
\begin{align*}
  h_1 &= f(x_{k - 1}) & h_2 &= f(x_{k - 1} + \deltat/2 h_1)  \\
  h_3 &= f(x_{k - 1} + \deltat/2 h_2) & h_4 &= f(x_{k - 1} + \deltat/2 h_3)
\end{align*}
for $k = 1, \dots, K$.
For leapfrog integration we separate the components of the state $x = (q, p)$ and $f(q_k, p_k) = (\dot{q}_k, \dot{p}_k)$ and compute:
\begin{align*}
  p_{k+1/2} &= p_k + \frac\deltat2 \dot{p}_k\\
  q_{k+1} &= q_k + \dot{q}(q_k, p_{k+1/2}) \deltat\\
  p_{k+1} &= p_{k+1/2} + \frac\deltat2 \dot{p}(q_{k+1}, p_{k+1/2})
\end{align*}
where the notation $\dot{q}(q_k, p_{k+1/2})$ denotes the $\dot{q}$ component of $f(q_k, p_{k+1/2})$ and analogously for $\dot{p}$.

We also consider the implicit Euler method, which is given by the potentially nonlinear equation
\[
x_{k} - \deltat f(x_{k}) = x_{k - 1}
\]
that is solved in each time step $k = 1, \dots, K$.

We tested another implicit method, BDF2. This is a second order multistep method with the formula given by
\[
x_{k} - \frac{4}{3} x_{k - 1} + \frac{1}{3} x_{k - 2} = \frac{2}{3} \delta t f(t_{k}, x_{k})
\]
To kickstart this method, which requires two steps of history, we initially do one step of backward Euler. This maintains the stability and error properties of the method.

\section{Learning methods}
\label{sec:method-detail}

\subsection{Training}
\label{sec:training}

Training for both step and derivative problem formulations is done with the Adam \citep{kingma2014adam} optimizer for all neural networks, except the neural network kernel which uses standard stochastic gradient descent with learning rate $0.001$ and weight decay $0.0001$. With the Adam optimizer, no weight decay is used, and most networks use a learning rate of $1\times10^{-3}$. Exceptions to this are: CNNs, MLPs and the u-net for Navier-Stokes, and CNNs and MLPs on the spring mesh. For both of these systems the CNNs and MLPs use a learning rate of $1\times10^{-4}$ and the u-net uses $4\times10^{-4}$.

On the Navier-Stokes system we also perturbed each batch of training data with normally-distributed noise with a variance of $1\times10^{-3}$. For step prediction the previous step was corrupted and the subsequent step left uncorrupted. For derivative prediction, the derivatives were updated to correct for the noise (i.e. $\tilde{x} = x + \mathcal{N} \implies \tilde{\dot{x}} = \dot{x} - \mathcal{N}$ where $\mathcal{N}$ is the sampled noise). This is inspired by the approach taken in \citet{pfaff20} and we found it to improve stability for neural networks on the Navier-Stokes system.

The number of training epochs varies based on the target system. On spring, wave, and spring mesh the networks are trained for 400, 250, 800, and 800 epochs, respectively. When reporting evaluation errors below, we average errors over all time steps of each randomly-sampled trajectory in the held-out evaluation set.

We train three independent copies of each neural network. When evaluating these, each test trajectory is evaluated with each duplicate neural network and the performance results are collected and processed together. Variance in plots of these results is produced both by the differences in performance for the three duplicated neural networks, and differing performance across the sampled evaluation trajectories.

\subsection{KNN regressor}%
\label{sec:knn-regressor}

We use a $k$-nearest neighbors regressor to predict the value of the
state derivatives, using $k=1$. With this method
$\tilde{f}_{\theta}(\tilde{x}^{(i)}_{k})$ finds the closest matching point in the
training set, and uses that point's associated derivatives as its
approximation, $\tilde{\dot{x}}^{(i)}_{k}$ in the case of derivative prediction.
For direct step prediction, the KNN finds the closest point and returns the next time step from that point's trajectory in the training set.
We use the KNN implemented in scikit-learn \citep{sklearn}, along with its default Minkowski metric.

\subsection{Kernel methods}%
\label{sec:nn-kernel}

Kernel methods provide a nonparametric regression framework \citep{scholkopfsmola}. In this benchmark we consider dot-product kernels of the form
$k(x,x') = \eta( \langle x, x' \rangle)$, which can be efficiently implemented in their primal formulation 
using random feature expansions \citep{rahimirecht} via the representation
\[
k(x,x') = \mathbb{E}_{z \sim \nu} [\rho(\langle x, z \rangle) \rho(\langle x', z \rangle)] \approx \frac{1}{L} \sum\nolimits_{l=1}^L \rho(\langle x, z_l \rangle) \rho(\langle x', z_l \rangle)~,
\]
where $\nu$ is a rotationally-invariant probability distribution over parameters and $z_l \sim \nu \text{ iid}$. The resulting maps $x\mapsto \rho(\langle x ,z_l \rangle)$ are \emph{random features}, associated with a shallow neural network with `frozen' weights. While further choices of kernel may be considered in the future, dot-product kernels have flexible approximation properties and are easily scalable \cite{rudi_kernels}.

In our experiments, we use $\rho=\text{ReLU}$ and $L=32768$ random features and train using kernel ridge regression.
We do not apply this approach to our Navier-Stokes system as its large state dimension makes achieving a sufficiently large set of random features infeasible.

\subsection{Deep networks}%

\paragraph{MLPs} We apply simple multilayer perceptron (MLP) networks in a variety of sizes. The configuration of the MLPs used varies with the target system. 
In particular, we divide our two systems into two classes: those with smaller state dimension (the spring and wave systems), and those with a larger state dimension (the spring mesh, and the Navier-Stokes problem).
We describe these architectures in terms of ``depth'' and ``width.'' The depth denotes the number of fully-connected operations in the MLP, so that for a depth of $d$ there are $d-1$ hidden layers. The width is the size of each hidden layer; the input and output dimensions are fixed by the state dimension of the system. The MLPs use $\tanh$ activations.

For the small systems we use three MLP architectures: (1) a depth of 2 and a hidden dimension (width) of 2048, (2) a depth of 3 and width of 200, and (3) a depth of 5 and a hidden dimension of 2048. For the large systems, we use two architectures: (1) a depth of 4 and width of 4096, and (2) a depth of 5 and width of 2048. The $10\times10$ spring mesh merges both sets of MLP architectures.

For the Navier-Stokes and spring mesh systems, the MLP gets as input both the current network state, and a one-hot mask indicating which points in the discrete simulation space are ``fixed,'' meaning either a boundary point, a point in an obstacle, or an immovable, fixed particle. 

\paragraph{CNNs} We also test several feed-forward convolutional neural networks. These use $\mathrm{ReLU}$ activations and we specify their architectures by a kernel size, and internal channel count. We use these simple CNNs only on the larger systems: the spring mesh and the Navier-Stokes. For both of these systems we test two CNN architectures: both have a kernel size of $9\times9$ and, respectively, 32 and 64 channels internally. The number of input channels is fixed by the system. Both systems have five: for the spring mesh, two channels each for position and momentum; and for the Navier-Stokes system two channels for velocity, one for pressure field, and two more for one-hot masks highlighing boundaries and the obstacles.

\paragraph{U-net} Finally, we implement another convolutional network---only for the Navier-Stokes system---a u-net following the architecture tested in \citet{thuerey20}. That work applied this architecture to another Navier-Stokes problem, predicting a single step of flow about an airfoil profile. Here we adjust the input and output channels of this architecture, and test on our Navier-Stokes problem, performing several recurrent steps of derivative or step prediction around circular obstacles.

The architecture itself consists of seven convolution operations on both the downsampling and upsampling side. The convolutions have a mix of $4\times4$ and $2\times2$ kernels, and have strides of two. The network includes skip connections common to u-net-style architectures. With each downsampling, the number of channels is doubled starting from an internal channel count of 64. Our Navier-Stokes system has a grid size of $221\times42$. To accommodate the amount of downsampling in this architecture we first upsample to $256\times256$ with bilinear interpolation.

\subsection{Other experimental details}
Our experiments were conducted on NYU's research HPC system, Greene. Neural networks were predominantly trained using NVIDIA RTX8000 GPUs, with a few runs on V100 GPUs. CPU-based runs used Intel Xeon Platinum 8268 CPUs. Our neural networks required, on average, approximately two hours to train and we consumed in total approximately 1785 hours of GPU time, across all our experiments, including some early experimental and exploratory runs not discussed here. Our dataset generation and non-neural network evaluation runs, which do not use GPUs, consumed approximately 2270 core-hours of CPU time, again including some exploratory runs. Datasets were generated using CPUs only. Neural network training and evaluation passes ran using GPUs through PyTorch. Evaluations and trainings of baseline numerical integrators and KNNs ran on CPU only.

\section{Experiment results}
\label{sec:plots}

\begin{figure*}
   \centering
   \includegraphics[width=\textwidth]{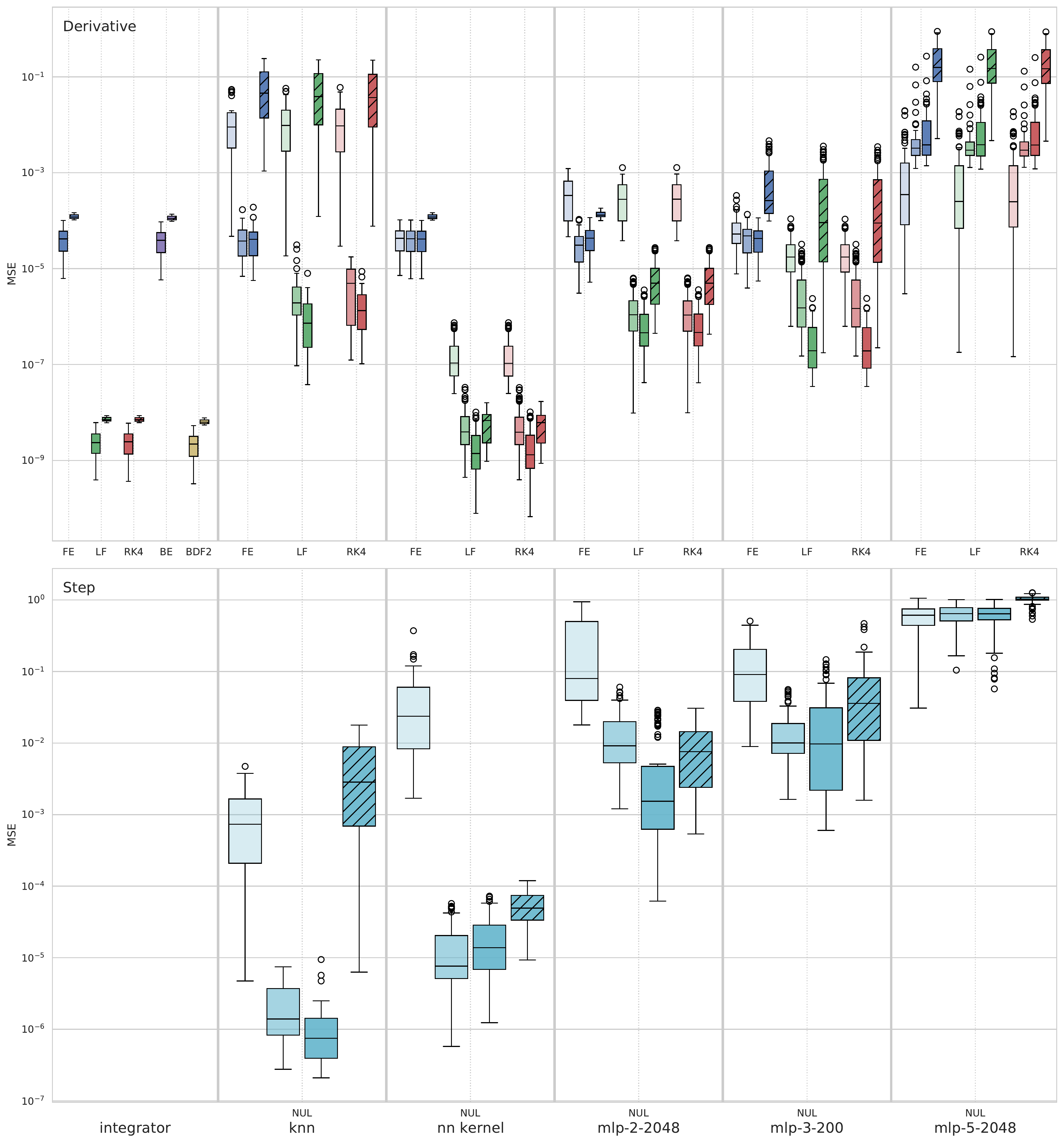}
   \caption{Error distribution for spring system for multiple training set sizes as well as out-of-distribution results.}%
   \label{fig:spring-error-dist}
 \end{figure*}

\begin{figure*}
  \centering
  \includegraphics[width=\textwidth]{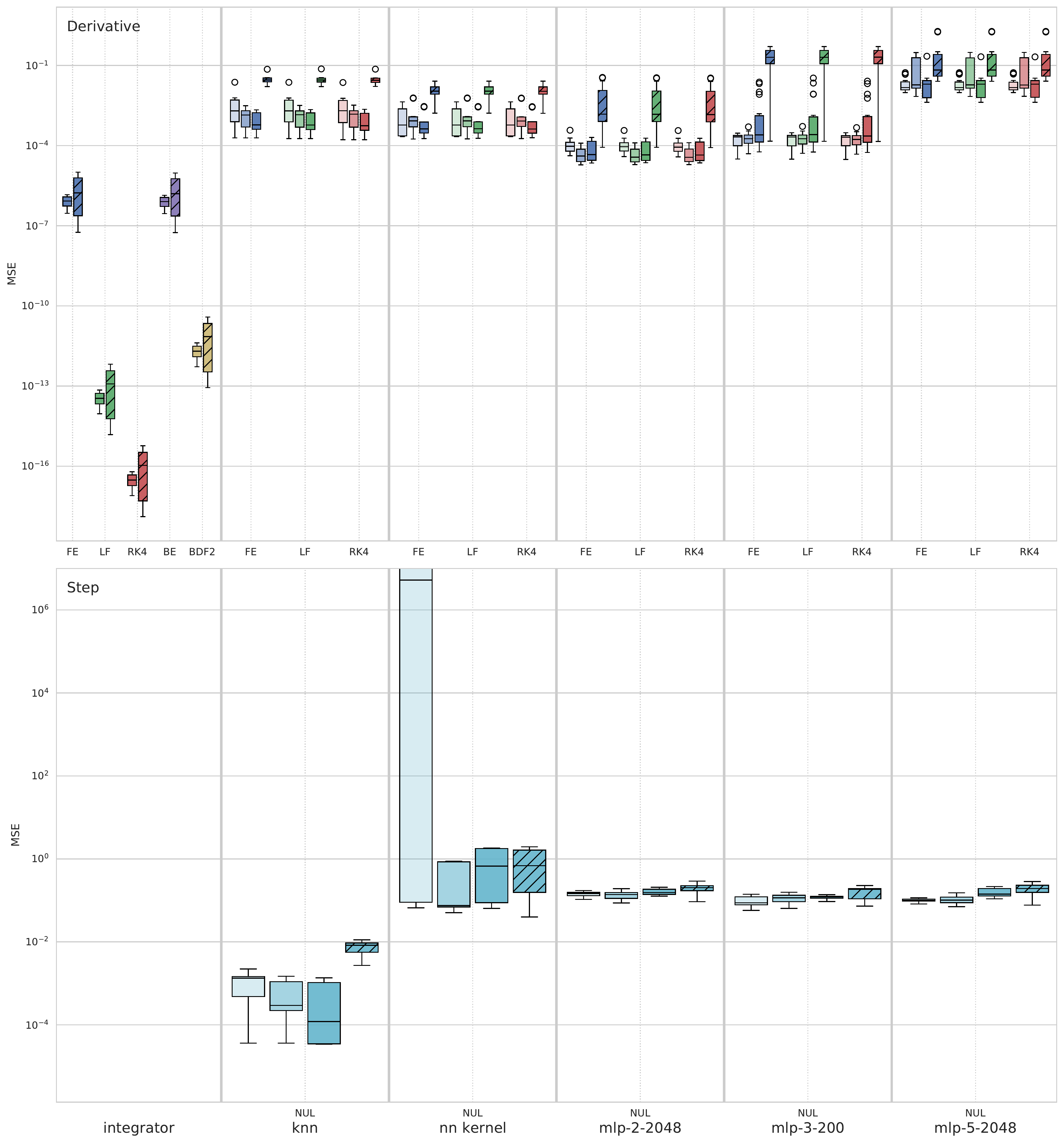}
  \caption{Error distribution for wave system for multiple training set sizes as well as out-of-distribution results.}%
  \label{fig:wave-error-dist}
\end{figure*}

\begin{figure*}
  \centering
  \includegraphics[width=\textwidth]{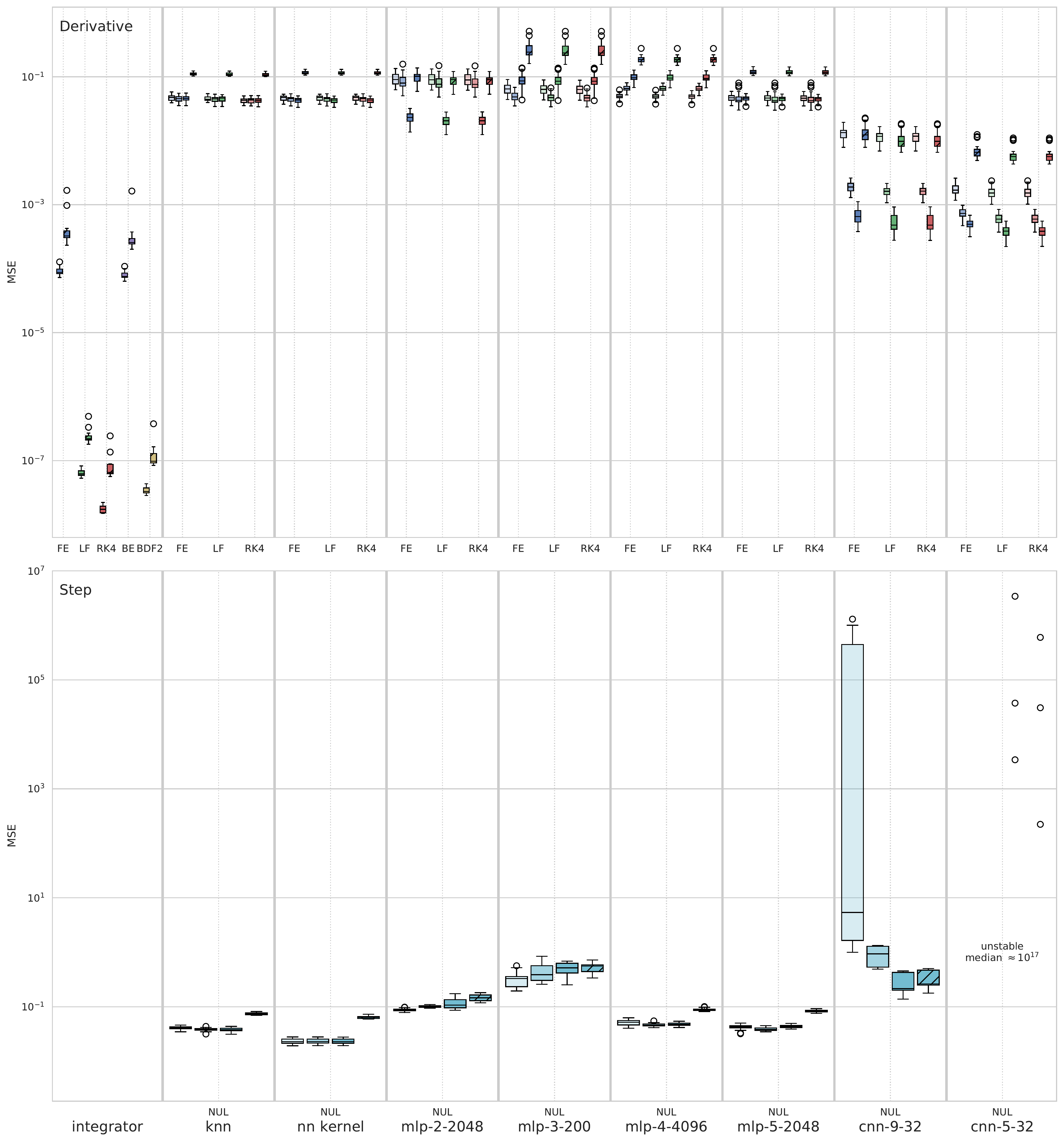}
  \caption{Error distribution for $10\times10$ spring mesh system for multiple training set sizes as well as out-of-distribution results.}%
  \label{fig:sm10-error-dist}
\end{figure*}

\begin{figure*}
  \centering
  \includegraphics[width=\textwidth]{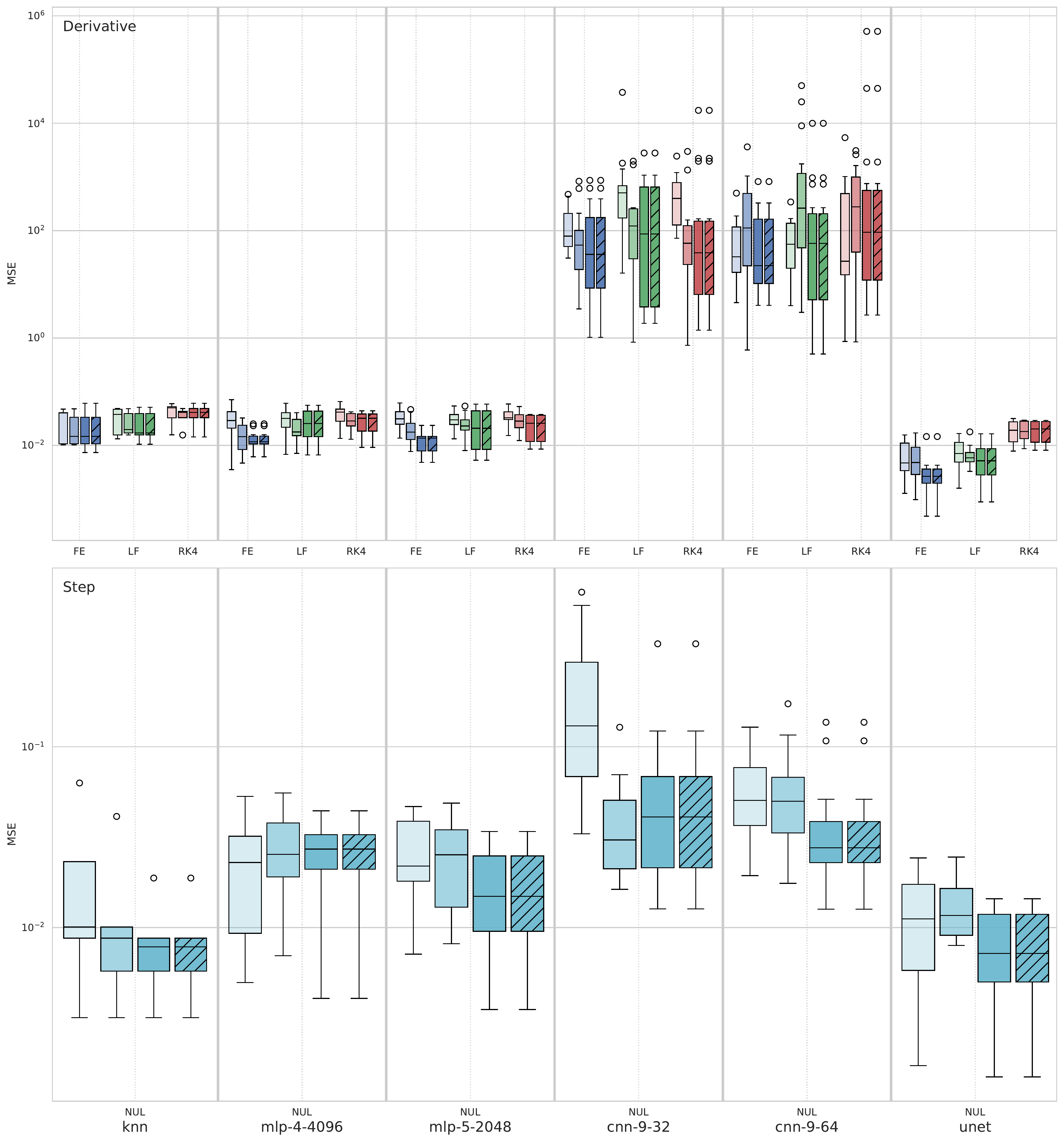}
  \caption{Error distribution for Navier-Stokes system for multiple training set sizes, and out-of-distribution results. Each trajectory has a single randomly-positioned obstacle. Note that this system does not have results for plain numerical integration.}%
  \label{fig:ns-single-error-dist}
\end{figure*}

\begin{figure*}
  \centering
  \includegraphics[width=\textwidth]{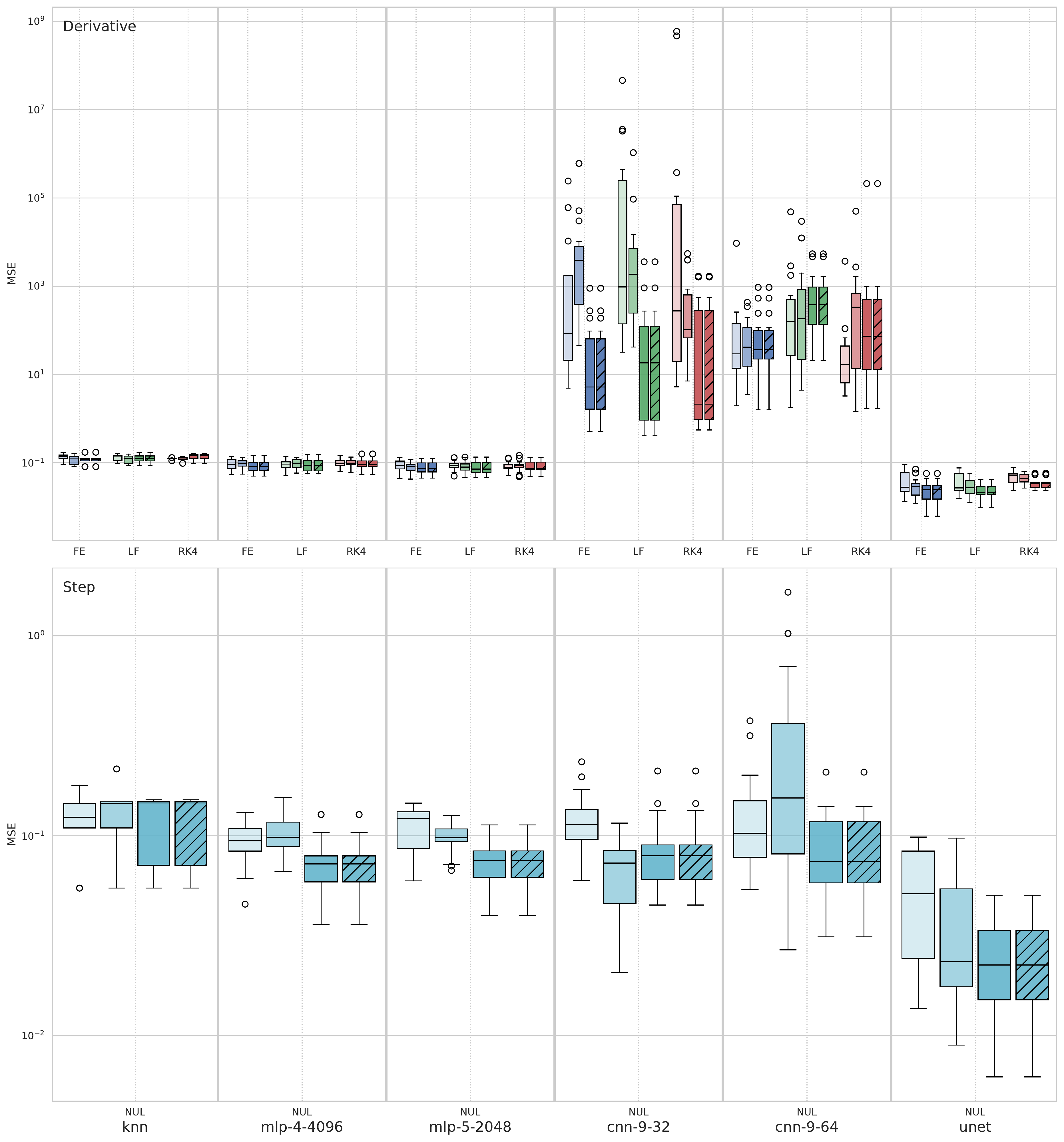}
  \caption{Error distribution for Navier-Stokes system for multiple training set sizes, and out-of-distribution results. Each trajectory has four randomly-positioned obstacles.}%
  \label{fig:ns-multi-error-dist}
\end{figure*}

To illustrate the error distribution for each neural network over the evaluation sampling distribution, we plot the errors as a box plot. Figures~\ref{fig:spring-error-dist}, \ref{fig:wave-error-dist}, \ref{fig:sm10-error-dist}, \ref{fig:ns-single-error-dist}, and \ref{fig:ns-multi-error-dist} show these error distributions, one plot for each system configuration.

Each plot is divided into two panes: one for derivative, and the other for step prediction. The datasets and training protocols followed are identical between the two task formulations. In each, the boxes are grouped first according to learning method, labeled at the bottom on the $x$-axis. For derivative prediction, the boxes are assembled into sub-groups according to the integrator applied (forward Euler/FE, leapfrog/LF, RK4, backward Euler/BE, or BDF2). These integrators are also indicated by the color of the box. In each group, from left to right the boxes become darker; this indicates the increasing training set size (see Table~\ref{tab:train-param}). The final box is hatched; this shows the evaluation results on the out-of-distribution set for the network exposed to the largest training set.

The boxes illustrate the distribution over per-trajectory average errors. For each system configuration (a system, derivative/step prediction, learning method, integrator, and particular training set size) we compute the per-step MSE against a ground truth result; these per-step errors are averaged to produce an error estimate for the trajectory. We also train three independent instantiations of each neural network architecture and evaluate each of these on all trajectories independently. These three repetitions of each trajectory for each network are included as part of the distribution in the box plot. The KNNs and numerical integrators are run a single time each. The errors of these different sampled trajectories form the distribution summarized by the box plot. The variance in the results is produced by a combination of the training results for the three copies of each network, and by the varying performance on each of the sampled evaluation trajectories. These plots were generated using Matplotlib's \cite{matplotlib} box plot routines. The box itself ends at the first and third quartiles of the data and the line in the middle is placed at the median of the data. The whiskers extend past the box by 1.5 times the size of the box. Circles are plotted for outlier points which lie outside the range of the whiskers. The plots here have a logarithmic $y$-axis to accommodate the wide range of error values, thus the boxes do not appear symmetric.

\FloatBarrier

\subsection{Weighted errors}
\label{sec:plots-weighted}

In most cases, due to accumulated errors, per-step errors increase as
numerical integration proceeds away from the initial condition. To
compensate for this trend and in an effort to explore the impact of
early vs.\ late step errors, we include several plots of error
distributions for which each time step's MSE has been weighted. To
produce these weights, each step's MSE is scaled by a value
$1/\exp(\ln(10^{2}) \cdot p_t)$ where $p_t \in [0, 1]$ is a scalar
representing the proportional time of the step (zero at start of the
trajectory, and one at the end). This produces an exponential decay
from the initial steps to the end and reduces the contribution of the
final steps by two orders of magnitude. These scaled MSEs are then
averaged for each trajectory and each neural network retraining as in
the plots above.

The results of these distributions for the Navier-Stokes system---both
single- and multi-obstacle forms---are included in
Figure~\ref{fig:ns-single-error-dist-weight} and
Figure~\ref{fig:ns-multi-error-dist-weight} below. A change in the
relative behavior of the learned methods is most visible in the step
prediction results in Figure~\ref{fig:ns-multi-error-dist-weight}.
Without the weighting, many of the learning methods perform comparably
to the KNN; however when emphasizing early steps, these methods
demonstrate improved errors relative to the re-weighted KNN errors.
This indicates that the learned methods outperform the accuracy of the
KNN on the early steps, but are somewhat unstable as the simulation
progresses.

For other systems, we did not observe significant changes in relative
performance of the learned methods. MSE distributions shifted, but
roughly in proportion to each other. This represents a greater general
stability in the learned methods on other systems, likely reflecting
the more predictable long-term behavior of the other systems. The
spring system is periodic, the wave system is stable over time, and
the spring mesh system has an energy decay term which simplifies and
stabilizes its long-term evolution. As a result, in most cases,
successfully learning the target task permits the learned methods to
maintain some stability over time, which decreases the relative effect
of the per-step weighting.

\begin{figure*}
  \centering
  \includegraphics[width=\textwidth]{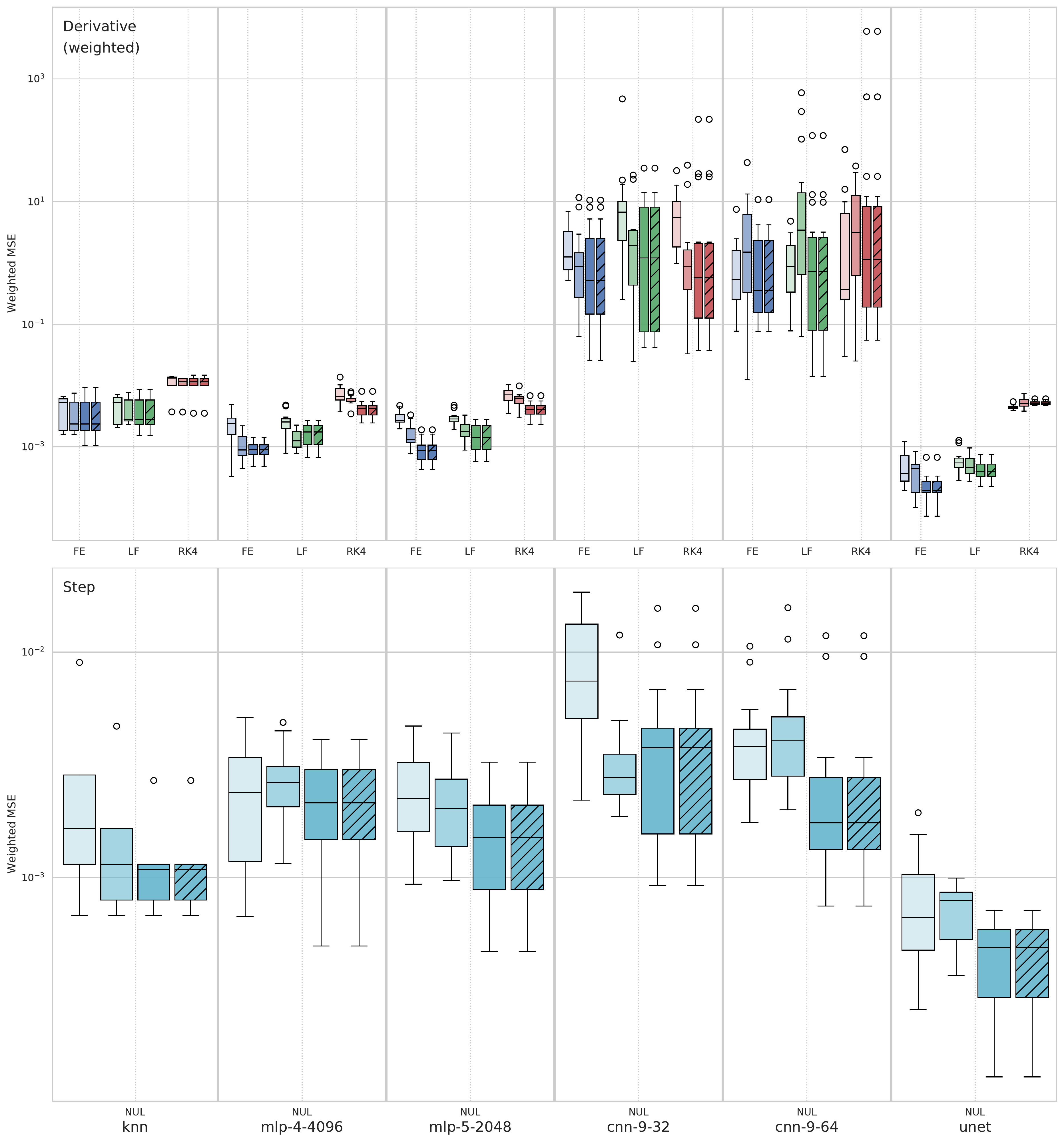}
  \caption{Error distribution for Navier-Stokes system for multiple training set sizes, and out-of-distribution results. Each trajectory has a single randomly-positioned obstacle. Per-step errors are weighted to decrease the contribution of later time steps with higher errors.}%
  \label{fig:ns-single-error-dist-weight}
\end{figure*}

\begin{figure*}
  \centering
  \includegraphics[width=\textwidth]{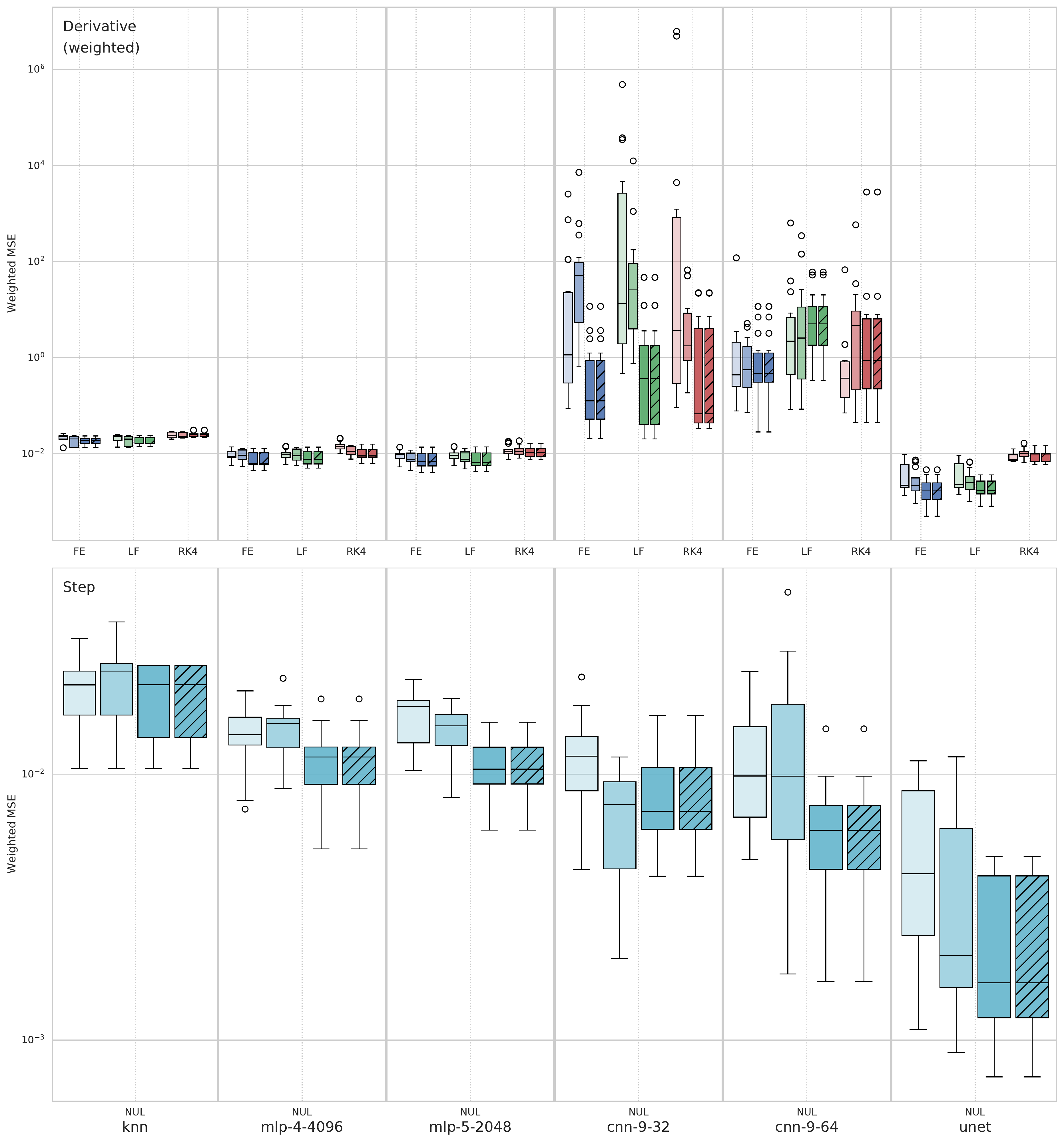}
  \caption{Error distribution for Navier-Stokes system for multiple training set sizes, and out-of-distribution results. Each trajectory has four randomly-positioned obstacles. Per-step errors are weighted to decrease the contribution of later time steps with higher errors.}%
  \label{fig:ns-multi-error-dist-weight}
\end{figure*}

\FloatBarrier
\section{Dataset and software documentation}
\label{sec:dataset-info}

This section contains information documenting the contents, structure,
and intended uses of the datasets used in this work.

\subsection{Overview}
The datasets used in this work consist of snapshots gathered from
numerical simulations of dynamical systems. These simulations were
carried out as part of this work and the software used to generate
them as well as stored outputs are made available for use and further
modification. The software source code is available under the MIT
license, and the stored data is available under a Creative Commons
Attribution 4.0 license (CC BY 4.0).

Simulations are carried out for four system types, described in
greater detail in the main work, above: \textbf{(1)} spring, \textbf{(2)} wave, \textbf{(3)}
spring mesh, and \textbf{(4)} Navier-Stokes. Each simulation's outputs are
intended to be used for developing and testing machine learning
methods for numerical simulations. They include snapshots of each
system's state across several data channels, as well as time
derivatives, either of which can be used as learning targets.

An archival copy of the stored data and software source code has been placed
in the NYU
Faculty Digital Archive (\url{https://archive.nyu.edu/handle/2451/63285}) for long-term storage.
The source code is also available on GitHub at
\url{https://github.com/karlotness/nn-benchmark}.

The stored data for each simulation type is stored in two components:
a JSON file containing metadata for the particular simulation, and an
associated uncompressed NumPy \texttt{.npz}-formatted file containing
the numerical results. Details of the contents of these files are
provided below. Our experiments were carried out in Python and these
files are readable using the Python standard library's \texttt{json}
module and the widely-used NumPy library. For other languages or
environments, the \texttt{.npz} files are ZIP archives containing
NumPy \texttt{.npy} files whose format is documented by NumPy
\url{https://numpy.org/doc/stable/reference/generated/numpy.lib.format.html}.

\subsection{Stored format}

Each dataset is a directory storing two files: ``system\_meta.json''
and ``trajectories.npz''. The \texttt{.npz} file contains several
NumPy array records with various shapes and data types. The names of
each of these records are referenced in the JSON file (documented
below). When loading the data from these systems these names should be
treated as opaque and always sourced from the JSON-formatted metadata.
In some cases, the same name is referenced several times for purposes
of data deduplication. The \texttt{.npz} file is used for bulk storage
of numerical data, separated from general metadata.

The simulation snapshots are divided into trajectories, each defined
by a particular initial condition from which a series of snapshots is
taken at several later time steps, by numerical simulation. Each
trajectory is divided into several ``channels'' of data, in particular
separating various state quantities, state time derivatives, and masks
marking special spatial locations for that trajectory, particular to
that system. The JSON file also contains trajectory-level parameter
information, and settings for global system-level parameters.

The stored data sets are intended to be used to test against the same
snapshots used in this work, without needing to configure the
dependencies necessary to generate the snapshots. The process
of running the simulations and reproducing the tests described in this
work is discussed in a separate section, below.

\subsubsection{Top-level object contents}

The JSON-formatted metadata file contains important information used
when loading these data sets. The contents of certain sections vary by
simulated system to reflect differences in relevant parameters and
other generic data. However, each has a similar global structure. Each
JSON file contains a top-level object with the same four keys:
\texttt{system}, \texttt{system\_args}, \texttt{metadata}, and
\texttt{trajectories}.

The value of \texttt{system} is a string identifying which system is
stored in that dataset. Its value will be one of: ``spring'',
``wave'', ``spring-mesh'', or ``navier-stokes''.

The value of \texttt{system\_args} is another object storing the
parameters which were passed to the simulation code to generate these
snapshots. These contents vary per system, but may be useful in
understanding what settings were configured for the simulators.

The \texttt{system\_args} object stores several key-value pairs which
vary by system reflecting different global parameters which apply to
every trajectory in the set. However it also always contains a key
\texttt{system\_args.trajectory\_defs} whose value is an array of
objects which contain parameters which may vary per-trajectory.

In addition to parameters in these objects which vary by system,
several are always present with the same meaning:
\texttt{num\_time\_steps}, \texttt{time\_step\_size}, and
\texttt{subsample}. The field \texttt{num\_time\_steps} is an integer
which sets the number of snapshots which are to be generated,
including the first snapshot which contains the initial condition. The
parameter \texttt{time\_step\_size} is a floating-point value which
sets the time difference between each stored snapshot. The physical effect
of the time step varies per system. The parameter \texttt{subsample}
is an integer ($1$ or larger) which allows generating the dataset on
a finer time grid than is reflected in the stored trajectory. Values
greater than $1$ cause the simulator to run at a time step of
$\mathtt{time\_step\_size}/\mathtt{subsample}$ for
$\mathtt{num\_time\_step} \cdot \mathtt{subsample}$ steps, and to discard
intermediate snapshots to produce an output at an un-subsampled
stride. This allows generating data sets at a higher simulation
quality while keeping the same end time and desired number of steps.

The \texttt{metadata} key stores an object with key-value pairs
providing system-dependent information on particular global
parameters.

The \texttt{trajectories} key contains an array of objects giving
information about the stored trajectory data. These are likely to be
the most useful when loading the snapshots as this object also
provides a mapping from each system's data channels to the array in
the dataset's ``trajectories.npz'' file.

The objects in the \texttt{trajectories} array each contain
system-dependent per-trajectory metadata which will be discussed
below, but as elsewhere several entries are always present. The first
is a \texttt{name} entry which gives a human-readable name for the
trajectory. The keys \texttt{num\_time\_steps} and
\texttt{time\_step\_size} have the same value and function as
discussed above. The \texttt{timing} entry contains an object with
information on timing of the data generation process. At present this
object has one entry of its own: \texttt{traj\_gen\_time}, which gives
the time to generate this trajectory measured in seconds.

Beyond these, each object in the \texttt{trajectories} array contains
a key \texttt{field\_keys} storing an object. This object has keys for
each data channel in this system whose values are strings giving the
name of a record stored in the ``trajectories.npz'' file. These
mappings are the best way to determine the correspondence between a
trajectory in the dataset and the stored bulk arrays which make up its
snapshot data. The names used for keys follow a general pattern,
usually prefixed with the name of the trajectory discussed above, but
they \emph{should} be treated as opaque and always sourced from the
JSON files. This mapping is in some cases used to reduce duplication
and some array records may be referenced multiple times.

Next, we discuss per-system variation in the overall structure listed
above, describing the metadata components and data channels which are
specific to each system.

\subsubsection{Spring}

This system is identified by a \texttt{system} entry with value
``spring''.

Under \texttt{system\_args} this system has no additional global
parameters, only per-trajectory parameters in the
\texttt{system\_args.trajectory\_defs} array. These per-trajectory
objects have a system-specific \texttt{initial\_condition} attribute
which is a sub-object with attributes \texttt{q} and \texttt{p}, both
of which are floating-point, giving the initial values for the position
and momentum of the simulated spring.

The \texttt{metadata} object contains one attribute, \texttt{n\_grid}
for consistency with the wave system, below. Its value is always the
integer 1.

The objects in the \texttt{trajectories} array contain only the
standard values defining the number of time steps and the
\texttt{name} of each trajectory. The object \texttt{field\_keys}
defines name mappings for the data channels of this system described
below. Some additional details for each channel are included in
Table~\ref{tab:doc-spring-chans}.

The channels ``q'' and ``p'' give the position and momentum of the
spring, respectively. Both values evolve in the one-dimensional space
of the system. Channels ``dqdt'' and ``dpdt'' are time derivatives of
these two quantities. The vector ``t'' gives the time of each snapshot
in the trajectory.

\begin{table}[H]
  \centering
  \caption{Data channels for the Spring system. $N_t$
    denotes the number of time steps in the trajectory.}
  \label{tab:doc-spring-chans}
  \begin{tabular}{lllp{3.5cm}}
    \toprule
    Channel name & Shape      & Data type \\
    \midrule
    q            & $(N_t, 1)$ & float64   \\
    p            & $(N_t, 1)$ & float64   \\
    dqdt         & $(N_t, 1)$ & float64   \\
    dpdt         & $(N_t, 1 )$ & float64   \\
    t            & $(N_t)$    & float64   \\
    \bottomrule
  \end{tabular}
\end{table}

\subsubsection{Wave}

This system is identified by a \texttt{system} entry with value
``wave''.

Under \texttt{system\_args}, the wave system has two global
parameters: \texttt{n\_grid} and \texttt{space\_max}. The parameter
\texttt{n\_grid} controls how many points are sampled on a regular,
one-dimensional spatial grid covering the interval
$[0, \mathtt{space\_max}]$, with a periodic boundary condition.
\texttt{space\_max} is a floating-point value controlling the end
point of this interval.

System-specific per-trajectory elements in
\texttt{system\_args.trajectory\_defs} are \texttt{wave\_speed},
\texttt{start\_type}, and \texttt{start\_type\_args}. The
\texttt{wave\_speed} parameter is a floating-point value controlling
the distance the wave pulses travel in a unit of time.

The other two parameters control the shape and position of the initial
pulse. \texttt{start\_type} is a string parameter selecting the type
of pulse to form; at present only ``cubic\_splines'' is supported. The
value of \texttt{start\_type\_args} is an object with additional
parameters which affect the starting pulse. For a cubic spline these
are: \texttt{height}, \texttt{width}, and \texttt{position}. Each of
these is a floating-point value which scales the height and width of
the pulse, and selects its center point in the spatial interval.

The \texttt{metadata} object repeats the values of the
\texttt{n\_grid} and \texttt{space\_max} attributes described above.

Each object in the \texttt{trajectories} array contains the
trajectory's value of the \texttt{wave\_speed} parameter discussed
above. There are no further system-specific entries in these objects,
other than the data channels. Some attributes of these are given in
Table~\ref{tab:doc-wave-chans}.

The channels ``q'' and ``p'' give the position/height and vertical
velocity of the wave at each grid point, respectively. The channels
``dqdt'' and ``dpdt'' are time derivatives of these quantities. The
vector ``t'' gives the simulation time for each snapshot in the
trajectory.

\begin{table}[H]
  \centering
  \caption{Data channels for the Wave system. $N_t$
    denotes the number of time steps in the trajectory, and $N_p$ denotes
    the number of points in the spatial grid, determined by \texttt{n\_grid}.}
  \label{tab:doc-wave-chans}  
  \begin{tabular}{lllp{3.5cm}}
    \toprule
    Channel name & Shape        & Data type \\
    \midrule
    q            & $(N_t, N_p)$ & float64   \\
    p            & $(N_t, N_p)$ & float64   \\
    dqdt         & $(N_t, N_p)$ & float64   \\
    dpdt         & $(N_t, N_p)$ & float64   \\
    t            & $(N_t)$      & float64   \\
    \bottomrule
  \end{tabular}
\end{table}

\subsubsection{Spring mesh}

This system is identified by a \texttt{system} entry with value
``spring-mesh''.

Under \texttt{system\_args}, this system has a global parameter
\texttt{vel\_decay} which is a floating-point value configuring the
damping applied to the velocity of each mass. Per-trajectory elements
in \texttt{system\_args.trajectory\_defs} objects are
\texttt{particles} and \texttt{springs}, which are both arrays of
objects.

Each object in \texttt{particles} has three attributes:
\texttt{is\_fixed}, a boolean indicating whether this particle is fixed
in place and immovable; \texttt{mass}, a floating-point value for the
particle's mass; and \texttt{position}, an array of two floating-point
values giving the $x$ and $y$ coordinates of the particle's initial
position. The order of the particles is significant and their index in
this array is their index referenced in \texttt{springs}, below.

The objects in the \texttt{springs} array have four attributes:
\texttt{a}, \texttt{b}, \texttt{rest\_length}, and \texttt{spring\_const}.
The values of \texttt{a} and \texttt{b} are integers specifying which
two particles this spring connects in the order of the objects in the
\texttt{particles} array. The springs are undirected so the order of
\texttt{a} and \texttt{b} is not important. The values
\texttt{rest\_length} and \texttt{spring\_const} are floats giving the
rest length of the spring and its spring constant, respectively.
In this work, the edges described in the array are
one-hop nearest neighbors in each axis-aligned direction, and are the edges of
the regular, square grid.

The \texttt{metadata} object has attributes repeating some values from
\texttt{system\_args}: \texttt{edges} is an array of objects repeating
\texttt{springs} as described above, and \texttt{particles} repeats
the corresponding array, both sourced from the first entry in
\texttt{system\_args.trajectory\_defs}.

Beyond these in \texttt{metadata} are: \texttt{n\_dim}, giving the
spatial dimensions in which the particles move (in this work, always
2); \texttt{n\_grid}, repeating the same value; \texttt{n\_particles},
giving the length of the \texttt{particles} array; and the
\texttt{vel\_decay} value repeated here as well.

There are no system-specific entries in the objects in the
\texttt{trajectories} array, other than the data channels. Some
attributes of each of these are given in Table~\ref{tab:doc-sm-chans}.

The channel ``q'' and ``p'' give the per-particle position and
momentum, respectively. These values are provided in both $x$ and $y$
components for the two-dimensional space. The channels ``dqdt'' and
``dpdt'' give time derivatives for these quantities.

The channel ``t'' provides the simulation time at which each snapshot was taken.

The ``edge\_indices'' gives the locations of the edges (the springs)
between each particle. The integers in this channel index in the same
order as the per-particle $N_p$ dimension in the other channels. The
values in this channel are directed so each spring is repeated twice,
once with its two end indices in both orders.

The array for ``masses'' gives the mass of each particle in the same
order as the $N_p$ dimensions in other channels.

The channel ``fixed\_mask'' is a boolean mask with \texttt{true} for each particle which is fixed in place. The channels ``fixed\_mask\_q'' and ``fixed\_mask\_p'' are the same, except with repeated values to be suitable for broadcasting.

\begin{table}[H]
  \centering
  \caption{Data channels for the Spring-Mesh system. $N_t$ denotes the number of time steps in the trajectory, $N_p$
    denotes the number of particles, $N_e$ denotes the number of
    edges.}
  \label{tab:doc-sm-chans}  
  \begin{tabular}{lllp{3.5cm}}
    \toprule
    Channel name        & Shape           & Data type & Notes                  \\
    \midrule
    q                   & $(N_t, N_p, 2)$ & float64   &                        \\
    p                   & $(N_t, N_p, 2)$ & float64   &                        \\
    dqdt                & $(N_t, N_p, 2)$ & float64   &                        \\
    dpdt                & $(N_t, N_p, 2)$ & float64   &                        \\
    t                   & $(N_t)$         & float64   &                        \\
    edge\_indices       & $(2, N_e)$      & int64     &                        \\
    masses              & $(N_p)$         & float64   &                        \\
    fixed\_mask         & $(N_p)$         & bool      &                        \\
    fixed\_mask\_q      & $(N_p, 2)$      & bool      &                        \\
    fixed\_mask\_p      & $(N_p, 2)$      & bool      & Alias: ``fixed\_mask\_q'' \\
    extra\_fixed\_mask\ & $(N_p)$         & bool      & Alias: ``fixed\_mask'' \\
    \bottomrule
  \end{tabular}
\end{table}

\subsubsection{Navier-Stokes}

The Navier-Stokes system is identified by a \texttt{system} entry with
value ``navier-stokes''.

Under \texttt{system\_args}, this system has a global parameter
\texttt{grid\_resolution} which is a floating-point value, giving the
space stride of the regular grid on which the solutions are sampled.
Per-trajectory elements in \texttt{system\_args.trajectory\_defs}
objects are \texttt{viscosity} and \texttt{in\_velocity}, which are parameters
passed to the FEM solver giving the viscosity of the fluid and the
velocity of the incoming flow, respectively. Both are floating-point values.

The objects in \texttt{system\_args.trajectory\_defs} also have
a parameter \texttt{mesh} which describes the location of obstacles in
the simulated space. Each entry in the array is an object with two
keys: \texttt{radius}, a single floating point value for the radius of
the circular obstacle, and \texttt{center}, an array of two floating
point values giving the $x$ and $y$ position of the center of the
circle. These values allow placing multiple obstacles in the
simulation space and impact the mesh which is generated and used by
the external finite element solver.

The \texttt{metadata} entry for this system contains two values:
\texttt{grid\_resolution} and \texttt{viscosity}, a copy of the same
values as described above. The viscosity value is taken from the first
trajectory entry.

The objects in the \texttt{trajectories} array have extra global
parameter values: \texttt{in\_velocity}, and \texttt{viscosity}, which
are as discussed above. Their \texttt{field\_keys} entries have mapped
names for the data channels listed in Table~\ref{tab:doc-ns-chans}.
The ``q''- and ``p''-related channels are present as aliases for
consistency with other systems.

The core values for this system are the ``solutions'' and
``pressures'' channels which store the flow velocity of the fluid and
the pressure field, respectively. The two channels for the solutions
are the $x$ and $y$ flow velocities.

Separately the ``grads'' and ``pressures\_grads'' channels store
approximated time derivatives computed from neighboring time steps
from the FEM solver's output.

The ``t'' channel is a vector giving the simulation time of each snapshot.

The channels ``vertices'' and ``edge\_indices'' identify the spatial
position and neighboring grid points for each sample point,
respectively. ``vertices'' gives the $x$- and $y$-coordinates as
separate channels, and ``edge\_indices'' stores indexes into the
per-particle dimension $N_p$ of each spatial value. The edges
described in the array are one-hop nearest neighbors in each
axis-aligned direction, the edges of the regular, square grid.

The ``fixed\_mask'' channels are boolean masks for the sample points,
indicating which of them form part of the boundary or an obstacle.
``fixed\_mask'' itself stores a value of \texttt{true} for points
which are either a boundary or an obstacle. The arrays
``fixed\_mask\_solutions'' and ``fixed\_mask\_pressures'' store the
same, just repeated to match the dimensions of the corresponding data
channels, suitable for broadcasting and masking or other purposes. The
``extra\_fixed\_mask'' is like ``fixed\_mask'' except that it provides
boolean per-particle masks for \emph{two} kinds of points. The last
dimension of this mask separates the two sub-channels, one for each
type of mask. Mask $0$ has \texttt{true} for the obstacles and the
boundaries, while mask $1$ has \texttt{true} only for the obstacles.

\begin{table}[H]
  \centering
  \caption{Data channels for Navier-Stokes system. $N_t$
    denotes the number of time steps in the trajectory, and $N_p$
    denotes the number of points in the regular grid. $N_p=9282$ for
    the datasets in this work.}
  \label{tab:doc-ns-chans}  
  \begin{tabular}{lllp{4.75cm}}
    \toprule
    Channel name           & Shape           & Data type & Notes                             \\
    \midrule
    solutions              & $(N_t, N_p, 2)$ & float64   &                                   \\
    pressures              & $(N_t, N_p)$    & float64   &                                   \\
    grads                  & $(N_t, N_p, 2)$ & float64   &                                   \\
    pressures\_grads       & $(N_t, N_p)$    & float64   &                                   \\
    t                      & $(N_t)$         & float64   &                                   \\
    q                      & $(N_t, N_p)$    & float64   & Alias: ``pressures''              \\
    p                      & $(N_t, N_p, 2)$ & float64   & Alias: ``solutions''              \\
    dqdt                   & $(N_t, N_p)$    & float64   & Alias: ``pressures\_grads''       \\
    dpdt                   & $(N_t, N_p, 2)$ & float64   & Alias: ``grads''                  \\
    edge\_indices          & $(2, N_p)$      & int64     &                                   \\
    vertices               & $(N_p, 2)$      & float64   &                                   \\
    fixed\_mask            & $(N_p)$         & bool      &                                   \\
    fixed\_mask\_solutions & $(N_p, 2)$      & bool      &                                   \\
    fixed\_mask\_pressures & $(N_p)$         & bool      &                                   \\
    fixed\_mask\_q         & $(N_p)$         & bool      & Alias: ``fixed\_mask\_pressures'' \\
    fixed\_mask\_p         & $(N_p, 2)$      & bool      & Alias: ``fixed\_mask\_solutions'' \\
    extra\_fixed\_mask\    & $(N_p, 2)$      & bool      &                                   \\
    \bottomrule
  \end{tabular}
\end{table}

\subsection{Data generation}

The section above discussed how to make use of the stored data sets.
Here we document the process used to configure these data sets and
invoke the simulators to produce the snapshots. The steps used here
cover the very similar process of running the neural network training
and evaluation phases. Following the steps here on the run
descriptions we have distributed allows recreating the experimental
setup used in the report above.

\subsubsection{Dependencies}

This section includes instructions for configuring the software
environment.

While the software we have produced for this work is available under
an open source license, the required dependencies are made available
under a variety of other licenses. These include some proprietary
components such as NVIDIA's CUDA libraries, and Intel's Math Kernel Library (MKL).
Review the licenses of the required dependencies before installing
or running.

\paragraph{Anaconda} The majority of software dependencies can be
installed using the Conda package management tool
(\url{https://docs.conda.io}). The root directory of our software
project contains an environment definition in the file
``environment.yml''. Using this file will create an environment
\texttt{nn-benchmark} containing the Python dependencies needed for
this project. If you are obtaining these dependencies from another
source, the contents of this file include the names of the packages
which will be required.

\paragraph{PolyFEM} In addition to the Python libraries needed for the
project, if you wish to \emph{generate} new Navier-Stokes
trajectories, you will also need a copy of PolyFEM. The source code
for this software can be obtained from the PolyFEM GitHub repository
(\url{https://github.com/polyfem/polyfem/}). This portion of the
project requires a copy of PolyFEM linked with Intel's Math Kernel
Library (MKL). To do this, locate the root directory of your MKL
installation and build PolyFEM from the root directory of its source
code with:
{
\small
\begin{verbatim}
mkdir build
cd build
MKLROOT=/path/to/mkl/root/ cmake .. -DPOLYSOLVE_WITH_PARDISO=ON -DPOLYFEM_NO_UI=ON
make
\end{verbatim}
}
This will produce a binary \texttt{PolyFEM\_bin}, required to produce
new Navier-Stokes trajectories. This binary must either be placed in
the directory from which you will run the simulation software, or you
should specify the parent directory of this binary in the environment
variable \texttt{POLYFEM\_BIN\_DIR} so that it can be located.

\paragraph{Singularity container (optional)}
It is possible to run the software directly in a manually-created
Anaconda environment. However, for convenience we include a build
definition file for a Singularity container (\url{https://singularity.hpcng.org}). \nocite{singularity}
Building a \texttt{.sif} container from this definition will produce
an environment suitable for running the software, including PolyFEM
and the required environment variables. Consult the Singularity
documentation for more information on building these containers in
your computing environment.

If you choose to use the Singularity container, either place it in the directory from which you will run the simulation software or provide the path to
the resulting \texttt{nn-benchmark.sif} file's parent directory in the environment
variable \texttt{SCRATCH}. The run management scripts (described
below) will look in this directory for the container and use it to run
jobs if it is found.

\subsubsection{Run descriptions}

The software used to generate the datasets, train networks, and
perform evaluations takes arguments from JSON files, specified on the
command line. This makes it possible to provide a large number of
arguments, to submit the jobs in batches, and to detect tasks that
have failed or that remain outstanding. The structure of these files
is relatively complex, so we provide additional tooling to assist in
generating them.

These utilities are located in the \texttt{src/run\_generators}
directory, in \texttt{utils.py}. Examples of their use are included in
the other Python scripts in that directory. These tools consist of
several object definitions which define jobs to run. These are divided
into three phases: data generation, network training, and evaluation.
To produce the descriptions of the desired jobs, one constructs Python
objects representing each of these, and calls their
\texttt{write\_description(dir)} methods. This method takes a single
argument: the root directory under which the job descriptions and the
resulting outputs will be stored. Create a new directory for each
experiment.

Each task takes an \texttt{Experiment} object as an argument; this
principally sets a prefix on the resulting file names, and records the
experiment name in the run description file. This separation is not
enforced and results of jobs combining different \texttt{Experiment}
objects can freely be mixed.

Datasets are generated by creating various \texttt{Dataset} subclass
objects. Each of these takes as an argument an
\texttt{InitialConditionSource} which provides the sampling of initial
conditions described above. Be aware that these objects cache the
initial conditions they have previously generated. This ensures that
larger datasets drawn from the same source are always strict supersets
of smaller datasets. The initial condition sources have parameters
which control the distribution from which samples are drawn, and the
datasets themselves have parameters controlling the simulations which
are carried out from these samples (such as the time step size, number
of steps, etc.). The \texttt{InitialConditionSource} objects do not
represent jobs and do not have \texttt{write\_description} methods.

Neural network training tasks are created by providing both an
\texttt{Experiment} object and two datasets: one for training, and one
for validation. The objects representing each dataset are provided as
constructor arguments to the objects representing each type of neural
network. Most networks have parameters which control their
architecture, choosing kernel sizes, hidden dimensions, etc.

Finally, evaluation run descriptions are generated by the
\texttt{NetworkEvaluation} object. This takes an \texttt{Experiment}
object, the object representing the network training task, an object
for the evaluation set to use, and the numerical integrator to combine
with the network. The exception to this is configuring runs for the
KNNs. These do not have a normal training phase and run entirely at
evaluation-time. These evaluation objects take an additional parameter
for their training dataset. When run, the job will load this training
set, fit the KNN, and then proceed with the rest of the regular evaluation
phase. \texttt{KNNPredictorOneshot} runs a KNN for step prediction and
\texttt{KNNRegressorOneshot} runs a KNN for derivative prediction with
the integrator specified as an argument to its constructor.

As an illustration of configuring jobs using these utilities we
provide the Python scripts used to generate the run descriptions for
the experiments discussed above. Be aware that running these scripts
will sample \emph{new} datasets from the distributions specified
above. Each script takes the name of the directory created for its
experiment as an argument and writes the run descriptions to that
directory in a \texttt{descr} directory, with three subdirectories,
one for each of the three phases.

\subsubsection{Launching jobs}

Once the run descriptions are generated, running the jobs is in large
part managed by the \texttt{manage\_runs.py} script. This script can
inspect the experiment directory to identify runs which are
outstanding, appear possibly incomplete, or whose description files
have been modified after the job was launched. The script also runs
the jobs from the description file, either serially, or in parallel by
submitting to a Slurm queue.

\paragraph{Scanning} Running \texttt{python manage\_runs.py scan <experiment directory>} will output information about the state of all jobs in that experiment. The script will indicate whether the jobs are yet to be run (outstanding), appear to be incomplete, or whether their descriptions were modified after the job launched. Jobs in one of the error states (incomplete or mismatched descriptions) can be deleted by adding the \texttt{-{}-delete=<mismatch or incomplete>} argument. This extra flag deletes the jobs in the specified state with \emph{no} further confirmation.

\paragraph{Launching} \texttt{python manage\_runs.py launch <experiment directory> <phase>} where phase is one of \texttt{data\_gen}, \texttt{train}, or \texttt{eval} will launch all outstanding jobs for the specified experiment and phase. Wait for all jobs from earlier phases to complete before beginning the next phase. By default the script will attempt to run the jobs locally in serial, but if the \texttt{sbatch} program is detected it will submit jobs to the Slurm queue instead. This selection can be overridden by passing one of \texttt{slurm} or \texttt{local} to the \texttt{-{}-launch\_type} argument. If the Singularity container is being used the script may output a warning that the ``nn-benchmark'' Anaconda environment is not loaded. This warning can be ignored as the container will provide the necessary environment.

\subsubsection{Recreating experiments}
We provide the JSON run descriptions for the experiments discussed
above. Once the software environment is configured, the
\texttt{manage\_runs.py} script can be used to launch copies of these
experiments.


\end{document}